\documentclass{article}
\usepackage{arxiv}
\usepackage{color,soul}

\usepackage[utf8]{inputenc} 
\usepackage[T1]{fontenc}    
\usepackage{hyperref}       
\usepackage{url}            
\usepackage{booktabs}       
\usepackage{amsfonts}       
\usepackage{nicefrac}       
\usepackage{microtype}      

\usepackage{amsmath,amsfonts,amssymb,graphicx}
\usepackage{natbib}
\usepackage{comment}

\title{Sparsity through evolutionary pruning prevents neuronal networks from overfitting}

\author{
  Richard C.~Gerum \\
  Biophysics Group, Department of Physics\\
  Friedrich Alexander University Erlangen-Nürnberg (FAU), Germany \\
    \And
  André Erpenbeck\\
The Raymond and Beverley Sackler Center for Computational Molecular and Materials Science,\\
School of Chemistry, Tel Aviv University (TAU),
Israel\\
  \And
  Patrick Krauss \\
  Experimental Otolaryngology,\\ Neuroscience Lab, University Hospital Erlangen, Germany \\
  Cognitive Computational Neuroscience Group\\at the Chair of English Philology and Linguistics, \\Friedrich-Alexander University Erlangen-Nürnberg (FAU), Germany\\
  Department of Otorhinolaryngology/Head and Neck Surgery, University of Groningen, \\ University Medical Center Groningen (UMCG), The Netherlands
  \And
  Achim Schilling \\
  Experimental Otolaryngology,\\ Neuroscience Lab, University Hospital Erlangen, Germany \\
  Cognitive Computational Neuroscience Group\\at the Chair of English Philology and Linguistics, \\Friedrich-Alexander University Erlangen-Nürnberg (FAU), Germany\\
}

\renewcommand{\cite}{\citep}
\renewcommand{\hl}{}

\begin{document}

\maketitle

\noindent\textbf{Corresponding author:} \\
Dr. Achim Schilling  \\ 
Neuroscience Group \\
Experimental Otolaryngology \\
Friedrich-Alexander University of Erlangen-N\"urnberg \\
Waldstrasse 1 \\
91054 Erlangen, Germany \\
Phone:  +49 9131 8543853 \\
E-Mail: achim.schilling@uk-erlangen.de \\ \\


\newpage

\begin{abstract}
Modern Machine learning techniques take advantage of the exponentially rising calculation power in new generation processor units. Thus, the number of parameters which are trained to resolve complex tasks was highly increased over the last decades. However, still the networks fail --- in contrast to our brain --- to develop general intelligence in the sense of being able to solve several complex tasks with only one network architecture. This could be the case because the brain is not a randomly initialized neural network, which has to be trained from scratch by simply investing a lot of calculation power, but has from birth some fixed hierarchical structure. To make progress in decoding the structural basis of biological neural networks we here chose a bottom-up approach, where we evolutionaryly trained small neural networks in performing a maze task. This simple maze task requires dynamical decision making with delayed rewards. We were able to show that during the evolutionary optimization random severance of connections lead to better generalization performance of the networks compared to fully connected networks. We conclude that sparsity is a central property of neural networks and should be considered for modern Machine learning approaches.

\end{abstract}

\newpage

\section*{Introduction}

Sparsity is a characteristic property of the wiring scheme of the human brain, which consists of about $8.6\times 10^{10}$ neurons \cite{herculano2009human}, interconnected by approximately $10^{15}$ synapses \cite{Sporns2005, Hagmann2008}. Thus, from almost $10^{22}$ theoretically possible synaptic connections, only one of 10 million possible connections is actually realized. This extremely sparse distribution of both neural connections and activity patterns is not a unique feature of the human brain \cite{Hagmann2008} but can also be found in other vertebrate species such as for example mice and rats  \cite{Perin2011, Oh2014, Kerr2005, Song2005}. Even evolutionary very old agents with a quite simple nervous system such as the nematode \textit{C. elegans} with only 302 neurons \cite{jarrell2012connectome} and over 7\,000 connections \cite{J.G.WhiteE.SouthgateJ.N.Thomson1986} show this sparsity. Besides the described sparsity of virtually all nervous systems, many biological neural networks also show small world properties \cite{Watts1998, Amaral2000, latora2001efficient, Bassett2006}, such as scale free connectivity patterns \cite{Perin2011, VandenHeuvel2017, Bullmore2006}. 

\hl{When dealing with the term sparsity in biology and machine learning, it is necessary to distinguish between three different forms of sparsity: "sparse representation", "input sparsity", and "model sparsity" \mbox{\cite{kafashan2016relating}}. 
Sparse representation means that only a small amount of neurons respond to certain stimuli. Thus, even a fully connected network can have the property of sparse representation. One famous but controversial example for sparse representation often called sparse-coding \mbox{\cite{zhang2011sparse}} is the so called "grandmother cell", being a vivid term for the idea that only a single neuron encodes for one highly complex concept \mbox{\cite{quiroga2008sparse,barwich2019value, rose1996some}}. In recent years much effort has been undertaken to achieve sparse coding of certain input stimuli to improve machine learning \mbox{\cite{pehlevan2014selectivity, olshausen2004sparse, babadi2014sparseness, dasgupta2017neural}} \mbox{\cite{, jin2018eeg, jiao2018sparse}} and on the other hand sparse coding was also investigated in biology \mbox{\cite{crochet2011synaptic, zaslaver2015hierarchical}}. 
Input sparsity, in contrast, means that the input patterns fed to the neural network are sparse. However, in this study we investigate the development of model sparsity in artificial neural networks, being the analogue to a sparse connectome in biology. For reasons of simplicity, we use from now on the term sparsity to refer to model sparsity in the biological sense, as well as in the context of machine learning.}

Sparsity in biology is the result of both, phylogenetic and ontogenetic adaptations. Even though, almost all species' immature nervous systems are already very sparse, this sparsity is even further increased during development and maturation of the agents' nervous systems \cite{low2006axon}. In fact, the infant human brain contains two times more synapses than the adult brain \cite{kolb2011brain}. Analogously, the immature nervous system of \textit{C. elegans} contains more synapses than the adult form \cite{oren2016sex}.

But pruning is not restricted to axons and synaptic connections. It even extends to the total number of neurons, which also decreases during development. For instance, the immature nervous system of \textit{C. elegans} initially consists of 308 neurons \cite{chalfie1984neuronal}, whereas the adult form contains only 302 neurons \cite{jarrell2012connectome}. And also in humans, the number of neurons decreases during development \cite{yeo2004early}. These ontogenetic changes are referred to as pruning \cite{paolicelli2011synaptic}, and it seems to be a universal phenomenon for all species from \textit{C. elegans} to humans. Furthermore, pruning is found to be mandatory for healthy development \cite{hong2016new}. In cases where normal synaptic pruning fails, this may even lead to disorders like schizophrenia \cite{boksa2012abnormal}.

Since sparse connectivity architectures are realized on all scales and in a vast number of agents of different complexity, it can be assumed that sparse connectivity is a general principle in neural information processing systems, leading to advantages compared to densely connected networks. One major advantage of sparse artificial deep neural networks used for image classification in comparison to fully connected networks, is the reduction of computational costs while at the same time boosting the ability to generalize \cite{Han2015, Anwar2017, Wen2016, mocanu2018scalable}. 

However, these pure feed forward network architectures show low biological plausibility as they neither have the ability of dealing with time series data, nor have any memory-like features. 
Efficient processing of time series data in artificial neural networks is a complex task with a bunch of limitations. The technique of training the neural networks by unfolding the data in time is computational expensive and time consuming and leads to effects such as vanishing or exploding gradients \cite{Hochreiter1998, Pascanu2012}, which have been partly overcome by the introduction of Long-Short-Term-Memory Networks \cite{Schmidhuber1997}. However, these networks are difficult to interpret from a biological point of view. 

To overcome these limitation, novel biologically inspired approaches for processing time series data were introduced called reservoir computing \cite{Verstraeten2007, Lukosevicius2012}. A so called reservoir of neurons with fixed (i.e.\ not adjusted by training) random recurrent connections is used to calculate higher-order correlations of the input signal which serve as input for a feed forward output layer that is trained with error back-propagation. The properties of the reservoir networks were found to be ideal for biologically inspired parameters with a high sparsity \cite{Alexandre2009}. Thus, these reservoirs work best at the edge of chaos, meaning that the parameters have to be chosen so that they are balanced between complete chaos and absolute periodicity \cite{Schrauwen2007, Bertschinger2004, Krauss2019}.

Much effort has be undertaken to apply the technique of reservoir computing on tasks with delayed rewards such as robot navigation in mazes  \cite{Antonelo2012, Antonelo2007}.

However, the technique of reservoir computing is still based on the fact that the output layer has to be trained in a supervised way using back-propagation and, thus, complex tasks with a delayed reward are difficult to realize. 
Thus, in this study we used an evolutionary approach to train networks in solving a maze task. \hl{Although, reinforcement learning techniques---especially deep reinforcement learning---would be suitable for this kind of hidden-state Markov process, this approach lacks biological plausibility. Thus, deep reinforcement learning needs further techniques for hyper-parameter tuning such as Bayesian optimization \mbox{\cite{springenberg2016bayesian}}. However, the aim of this study is to analyze self-evolving networks and to characterize the architecture to derive basic principles, which are of relevance for biological systems. Nevertheless, evolutionary techniques could be used to optimize reinforcement learning as well \mbox{\cite{young2015optimizing}}}.

In our evolutionary system we were able to show that the random severing of connections (evolutionary pruning), without explicitly rewarding sparsity, did lead to a general sparsification of the networks and a better generalization performance. \hl{Furthermore, we could demonstrate that evolutionary training works best when the probability for destroying connections is higher but in the same range of the probability for recreating connections. 

The paper is structured as follows: In the Method section we describe the used software resources, how the maze task is implemented, the evolutionary algorithm and the initial neural network architecture. The Results section starts with the analysis of the effect of the different sparsification methods on the performance of the neural networks. We furthermore analyze the architecture of the evolved networks and demonstrate the effect of severance probability and reconnection probability. We added a Conclusion section summarizing the main results, and provide a discussion on the limitations of the study and possible future research directions (Discussion).}

\begin{figure}[htbp]
\centering
\makebox[\textwidth][c]{\includegraphics{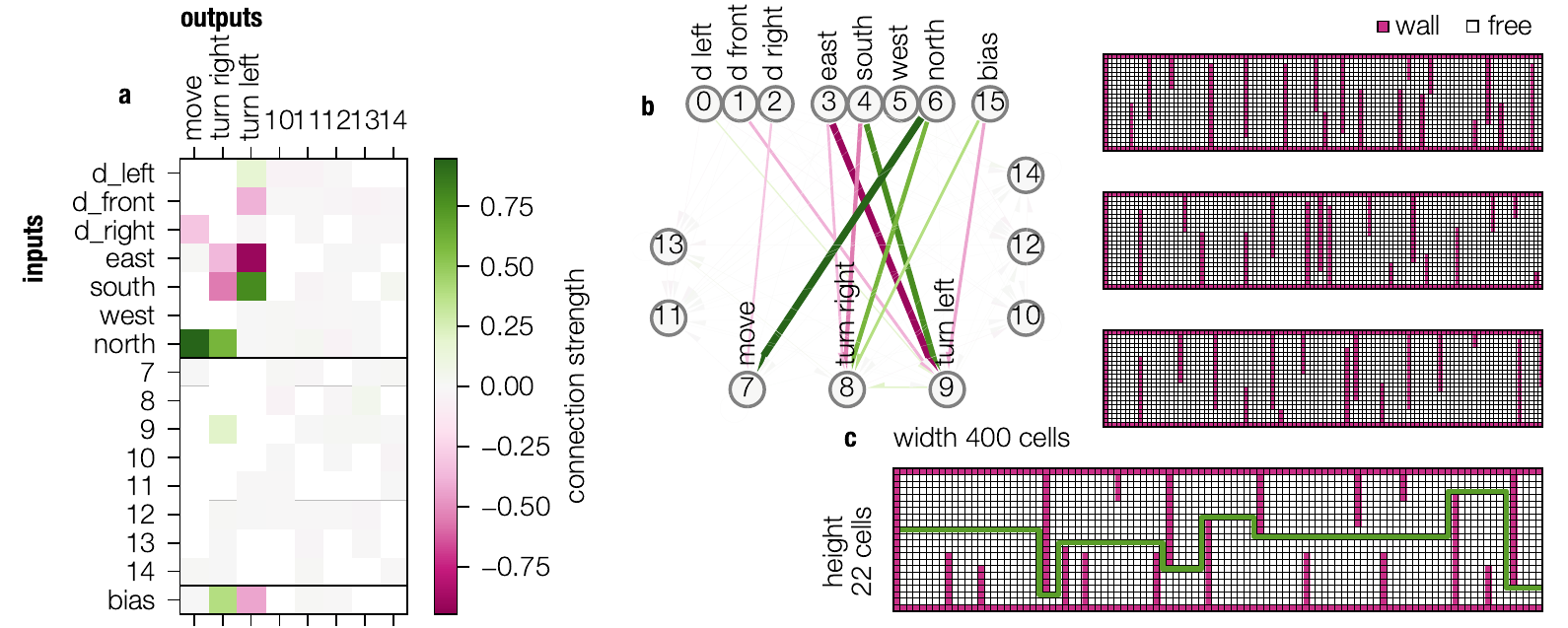}}
\caption{\textbf{Example network and mazes.} \textbf{a}, Weight matrix of an exemplary network. \textbf{b}, The same network displayed as connections. \textbf{c}, Mazes are 400 cells wide and 22 high. Walls (red) are at all borders and randomly placed in between. The green line depicts one ideal path though the network.}
\label{fig:Maze}
\end{figure}

\section*{Materials and Methods}

\subsection*{Software Resources}

All simulation were run on a desktop computer equipped with an i9 extreme processor (Intel) with 10 calculation cores. The complete software was written in Python 3.6 using the libraries sys, os, glob, subprocess, json, natsort, pickle, shutil, NumPy \cite{VanDerWalt2011b}. Data visualization was done by the use of Matplotlib \cite{Hunter2007a} and plots were arranged using the Pylustrator \cite{Gerum2019}.

\subsection*{Maze Task}

The task for the agents to perform is a maze based on a rectangular grid of 400x22 cells (Fig.\ \ref{fig:Maze}c). \hl{The maze/obstacle task we use here, is similar to the maze task described by Sanchez and coworkers \mbox{\cite{sanchez2001solving}}, which they also use to analyze the performance of evolutionary approaches.}

In our maze task there exist two types of cells, free cells and wall cells. Free cells can be entered and wall cells not. The border of the maze consists of walls to prevent agents from leaving the maze. Starting from the left, every 2 to 10 cells a wall with a length between 4 to 20 is inserted. With a probability of 0.25 the wall is inserted from the same side (up or down) as the last wall or with a probability of 0.75 it is inserted from the other side.

Agents start always at the left end of the maze facing to the right. As 'sensory' input each agent receives the distance to the wall in front, to the left and to the right  (input neurons 0 to 6, cf. Fig.\ \ref{fig:Maze}a).

If the distance is larger then 10, it is set to 10 (visual range). It also receives the direction, it is currently looking at, as a one-hot encoded, four-neuron input, i.e. one neuron at a time is in state 1 and the others are in state 0. This input serves as a kind of compass. The seven input neurons do exclusively receive input from the environment, but do not get any input from other neurons, thus, they are reset at each time step.

The agent can output three values for the three possible actions: go straight, turn right, or turn left. The action with the highest value is selected (winner takes all) (output neurons 7 to 9, cf. Fig.\ \ref{fig:Maze}a). When the agent chooses to go straight and the next field is a wall, it is not moved.

After 400 actions, the covered x-distance of the agent is fed to the fitness function which is proportional to the covered distance (cf. Fig.\ \ref{fig:Maze}c). Thus, the reward is delayed by 400 time steps. 

\subsection*{Network}

The logic of each agent consists of a fully connected network of $N=16$ neurons \hl{(identified as the number of neurons leading to the best performance with fast convergence, cf. Supplements {\ref{fig:plot_network_size}}, {\ref{fig:plot_network_size_small}}, {\ref{fig:plot_network_size_sparsity}})}, with states $s$ (cf. Fig.\ \ref{fig:Maze}a,b). The connection weights $W$ are initialized with a random value drawn from a uniformly distribution from the interval $[-\sigma, +\sigma]$ with $\sigma = 4\cdot\sqrt{\frac{6}{N+N}}/10$. For each time step $t$ in the task, the 3+4 input values (distances (left, front, right) and one-hot encoded direction) are set as the states of the neurons \#0 to \#6. Then one \hl{processing} pass \hl{through} the network is calculated, $s_{t+1} = \text{ReLu}\left(W\cdot s_{t}\right)$ \hl{($s_{t}$: state of the network at time point $t$)}, with $\text{ReLu}(x)$ being the rectified linear function 
\begin{align}
\text{ReLu}(x) =  \begin{cases} 
      0 & x\leq 0 \\
      x & x> 0\\
   \end{cases}
\end{align}
The states \hl{$s$} of the neurons \#7 to \#9 are used as outputs to choose one of the three possible actions to perform in the maze task (connectivity matrix see Fig.\ \ref{fig:Maze}a). Thus, the action is determined by a winner-takes all-method (in the case of no activation, the "move" action is chosen). Our approach is a policy based approach, as the output of the network directly is the action to take and no quality assessment of different states is undertaken.

\subsection*{Evolutionary Algorithm}

For optimizing the networks to fulfill the maze task, we use an evolutionary algorithm \cite{Fekiac2011}. Therefore, a pool of 1\,000 agents \hl{(for simulations with different population sizes cf. Supplements {\ref{fig:plot_population_size})}} is created with a random initialization and 10 mazes are created for the agents to be trained on.

For each iteration, all agents have to perform the maze task and are assigned a fitness, depending on their score in the task. The best half of the agent pool has now the chance to create offspring. The probability to generate an offspring is proportional to their relative fitness compared to the other agents. Agents with a probability of 10\% or more are set to a maximal probability of 10\% to retain biodiversity. For each of the old agents to be replaced, a parent agent is selected at random according to their reproduction probabilities. Agents can have multiple offspring or no offspring at all.

After offspring generation, each agent is mutated. We used three different mutation types:
\begin{itemize}
\item \textbf{Weight mutation}: The connection weights $W$ are each mutated by addition of a Gaussian distributed random variable ($\mu=0$\hl{: mean of distribution}, $\sigma = |\sigma_\mathrm{mut}|$\hl{: standard deviation}). 
\item \textbf{Mutation rate change}: The mutation rate $\sigma_\mathrm{mut}$ is also mutated by multiplication with a Gaussian distributed random variable ($\mu=1$, $\sigma = \sigma_\mathrm{mut}$). 
\item \textbf{Connection mutation}: Existing connections are removed with a probability $p_\mathrm{disconnect}$ and non-existing connections are added with a probability of $p_\mathrm{connect}=p_\mathrm{disconnect}$. Removed connections have a weight of 0 and are not subject to weight mutations. Thus, a removed connection cannot be recovered by a simple mutation step, but can only be recovered by a reconnection mutation. 
\end{itemize}
The fitness (eq.\ \ref{eq:fitness}) is calculated from the squared mean of the square root distances the agent reached in all 10 training mazes (SMR, eq.\ \ref{eq:smr}, this is done to favor generalizing agents which perform okay in all mazes against a specializing agent which performs well in only one maze), the maximum mean activation of the neurons and optionally from the mean number of active connections:
\begin{align}
\label{eq:fitness}
\mathrm{fitness} =& ~ \mathrm{SMR}_\mathrm{mazes} - \mathrm{max}_\mathrm{mazes}(\mathrm{mean}(\mathrm{activation})) \\
\notag
&+ f_\mathrm{sparsity} \cdot \mathrm{sparsity}\\
\label{eq:smr}
    \mathrm{SMR}_\mathrm{mazes} =& \left(\frac{1}{N_\mathrm{mazes}}\cdot\sum_{i=1}^{N_\mathrm{mazes}}\sqrt{d_i}\right)^2
\end{align}

\begin{table}
\centering
\caption{\textbf{Settings for the different experiments.} The initial mutation rate $\sigma_\mathrm{mut}$, the probability of removing or restoring a connection $p_\mathrm{connect}$ and the sparsity reward factor $f_\mathrm{sparsity}$.}
\begin{tabular}{l|lll}
\toprule
name & $\sigma_\mathrm{mut}$ & $p_\mathrm{connect}$ & $f_\mathrm{sparsity}$\\
\midrule
control & 0.01 & - & -\\
connection severance & 0.01 & 0.01 & - \\
connection severance & 0.01 & 0.001 & -\\
connection severance no mut & -& 0.01 & -\\
connection severance sparsity reward& 0.01  & 0.01 & 0.1\\
sparsity reward & 0.01 & - & 0.1\\
\bottomrule
\end{tabular}
\end{table}
\hl{(SMR$_{\mathrm{mazes}}$): Root-Mean-Square-Distance,  $d_i$: covered distance in maze $i$ , activation: average activation of neural network, N$_{\mathrm{mazes}}$: number of mazes, $f_{\mathrm{sparsity}}$: sparsity reward factor)}
All experiments are repeated for 5 different seeds of the random number generator that is used to obtain the initial weights and the mutations. The different repetitions were performed on the same 10 training mazes to keep them comparable. \hl{The performance of the networks is defined as the average distance covered in the mazes after 400 time-steps. The validation performance is the average distance in unseen validation mazes, whereas the training performs is the average distance in the ten known training mazes.}

\begin{figure}[htbp]
\centering
\makebox[\textwidth][c]{\includegraphics{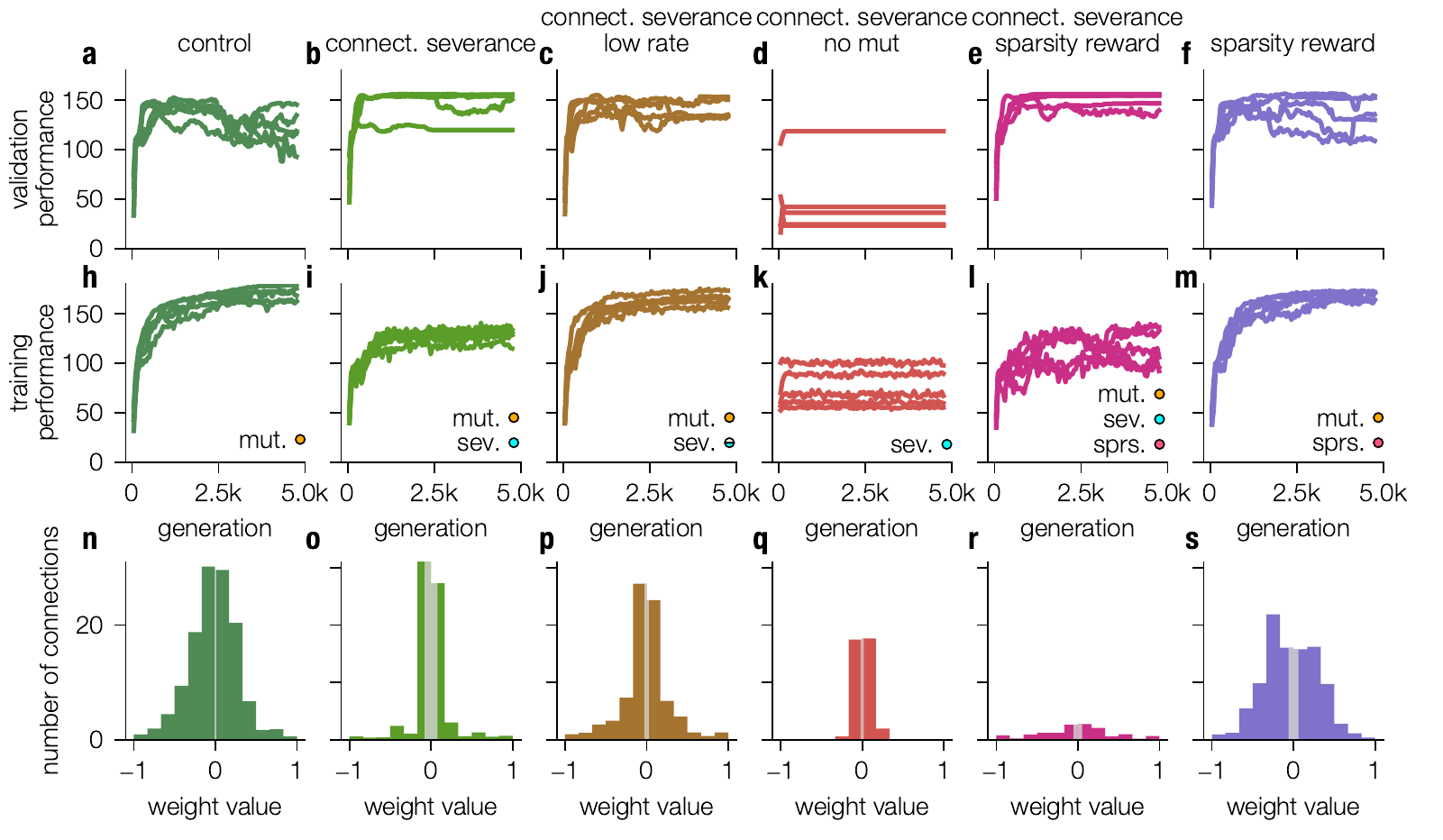}}%
\caption{\textbf{Performance of networks of the different experiments during training and validation.} \textbf{a-f}, Performance over generations for different experiments in 10 validation mazes. Curves show 5 different training seeds, evaluated on the same 10 validation mazes. \textbf{h-m}, Performance during training over generations for the 5 different training seeds. Labels stand for the different properties of the condition: "mut" for mutation of the weights, "sev" for severing/restoring connections, "sprs." for adding a sparsity reward to the fitness function. \textbf{n-s}, Histograms of the connection weights for the different experiments. Zero connections are not included in the histograms. Gray shaded areas in the center indicate which weights can be removed without reducing the performance.}
\label{fig:FitnessTrainValidation}
\end{figure}

\section*{Results}

The evolutionary algorithm was able to find solutions enabling the agents to efficiently navigate through the mazes. In all experiments, except the experiment without weight mutations, the agents gradually learned to perform better in the maze tasks over the generations (Fig.\ \ref{fig:FitnessTrainValidation}h-m). The convergence was quite slow, as about 5\,000 generations were needed to converge to a stable solution. The convergence behaviour was mostly independent of the seed of the random number generator, except for the "no mutation" condition, which relied strongly on the initial weights. The \hl{performance} during training was best for the conditions with no, or low sparsification pressure (Fig.\ \ref{fig:FitnessTrainValidation}h,j,m).

The fitness in validation mazes, that the agents had not seen during training, was more sensitive to the seed than the \hl{performance} during training. For the experiments with more sparsification pressure (Fig.\ \ref{fig:FitnessTrainValidation}b,e) the validation \hl{performance} did exceed the fitness during training, showing good generalisation, whereas the experiments with lower sparsification pressure (Fig. \ref{fig:FitnessTrainValidation}a,c,f) showed more problems with over-fitting, which means that the \hl{performance} is higher during training than during validation.

Validation \hl{performance} increased in most runs during training, although some drops in validation \hl{performance} were observed, which sometimes recovered after a few generations, but in some cases the validation \hl{performance} continued to fluctuate (see Supplements \ref{fig:ValidationFitnessContol}-\ref{fig:ValidationFitnessConSevSparsRew}). In general when the mean validation fitness dropped also the variability of the \hl{performance/fitness} increased, showing that even if some agents found an "overfitting" solution, it was not quickly adopted by all agents, whereas when the solutions were more general, they seemed more stable and were adopted by the whole pool of agents.

The weight distribution of the final generation was different for each experiment. While the experiments with low sparsification pressure, that also showed overfitting, show more connections and more weights with larger absolute values (Fig.\ \ref{fig:FitnessTrainValidation}n,p,s) compared to the other experiments that show more small weight values (Fig.\ \ref{fig:FitnessTrainValidation}o,q). Apart from "connection severance no mut" and "connection severance sparsity reward" (48\% and 49\% negative weights) all experiments show more negative weights (52--55\%), referring to inhibitory synapses, a fact which indicates more interesting behaviour and more efficient information processing \cite{Krauss2019}.

\begin{figure}[htbp]
\centering
\includegraphics{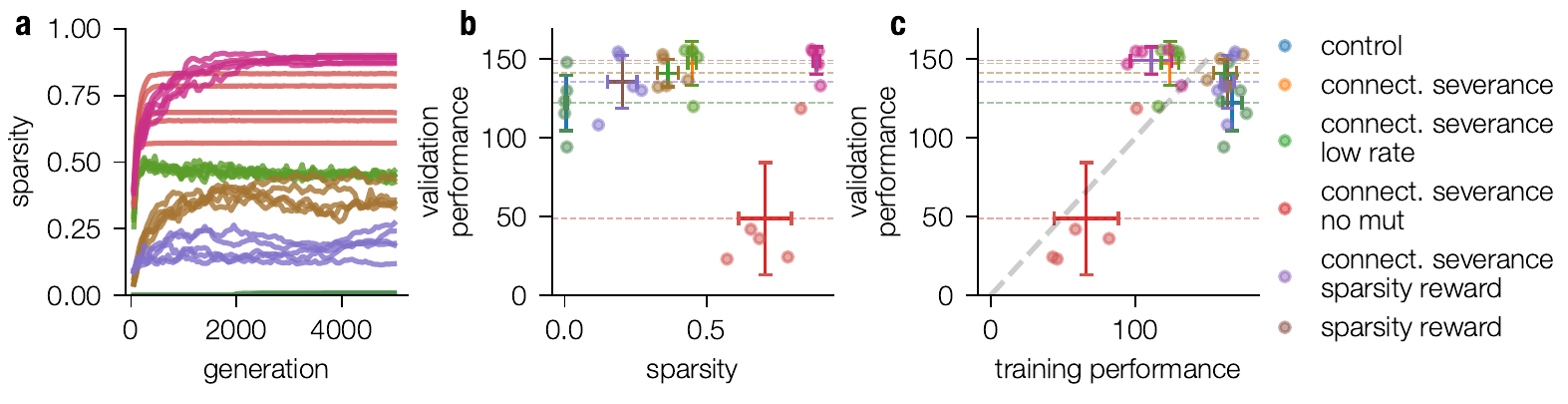}
\caption{\textbf{Correlation of validation performance to sparsity and training performance.} \textbf{a}, Sparsity ($1-\frac{n_{\text{non-zero}}}{n_{\text{possible}}}$) as a function of the epochs (generations, \hl{sparsity is exclusively gained by evolutionary pruning}). \textbf{b}, Correlation of validation performance to sparsity. Except for the case of "connection severance no mutation" validation performance increases on average with increased sparsity. \textbf{c}, Correlation of validation performance to training performance. Higher training performance leads in all experiments also to higher validation performance, indicating that none of the networks runs into severe overfitting.}
\label{fig:CorrelationSparsityValidation}
\end{figure}

In most experiments, the sparsity increased over time (Fig.\ \ref{fig:CorrelationSparsityValidation}a), but in "connection severance" it even slightly decreased after the initial rise and in "sparsity reward" the sparsity fluctuated strongly over time. A higher sparsity was in all cases (except the no mutation case) associated with a higher validation \hl{performance} (Fig.\ \ref{fig:CorrelationSparsityValidation}b), showing that sparsity improves the generalisation behaviour of evolutionary trained networks. A comparison of the training \hl{performance} to the validation \hl{performance} also shows that for the sparser cases, the training fitness decreases and in contrast to that, the validation \hl{performance} increases. Therefore, the sparsification prevents overfitting and enhances the generalisation \hl{properties of the networks}. In addition, it improves computation efficiency in the evolutionary trained networks as the computations, divided between many neurons in the fully connected control group, are, in sparser networks, forced to be carried out on a small subset of neurons.

Furthermore, it could be shown that the networks which perform best in the test mazes develop simple feed-forward structures (cf. Fig. \ref{fig:Maze}a).  Additionally, some asymmetry in the connectivity matrix can be observed (cf. Fig. \ref{fig:Maze}b). On the one hand, the bias units prefer the turn towards a certain direction. On the other hand, the connection from the input distance sensor to the turn output (e.g. to turn to the left side $d_\mathrm{left}$ in example Fig. \ref{fig:Maze}b) is over-represented for one side. Thus, the networks have a preference to go to one side (e.g. to turn left, in the example in Fig. \ref{fig:Maze}b), if there is enough space. The bias unit serves as counterpart, if the agent moves along the upper edge (resp. left side seen when moving along the x-axis) and guarantees that the network can walk away from the wall. The simple network architectures allow for the analysis of the functional tasks of certain neurons. This understanding of the functional tasks of neurons in artificial neural networks could potentially help to understand biological neural networks \cite{Jonas2017,Kriegeskorte2018}. Different experiments develop quite different solutions to solve the maze task (see Supplementary Material S1). Interestingly, in some experiments similar structures emerge, regardless of the seed. This is especially the case in the "connection severance" experiment, where 4 out of 5 solutions are strikingly similar. This hints at the existence of strongly attractive maxima in the space of possible solutions.

\begin{figure}[htbp]
\centering
\includegraphics{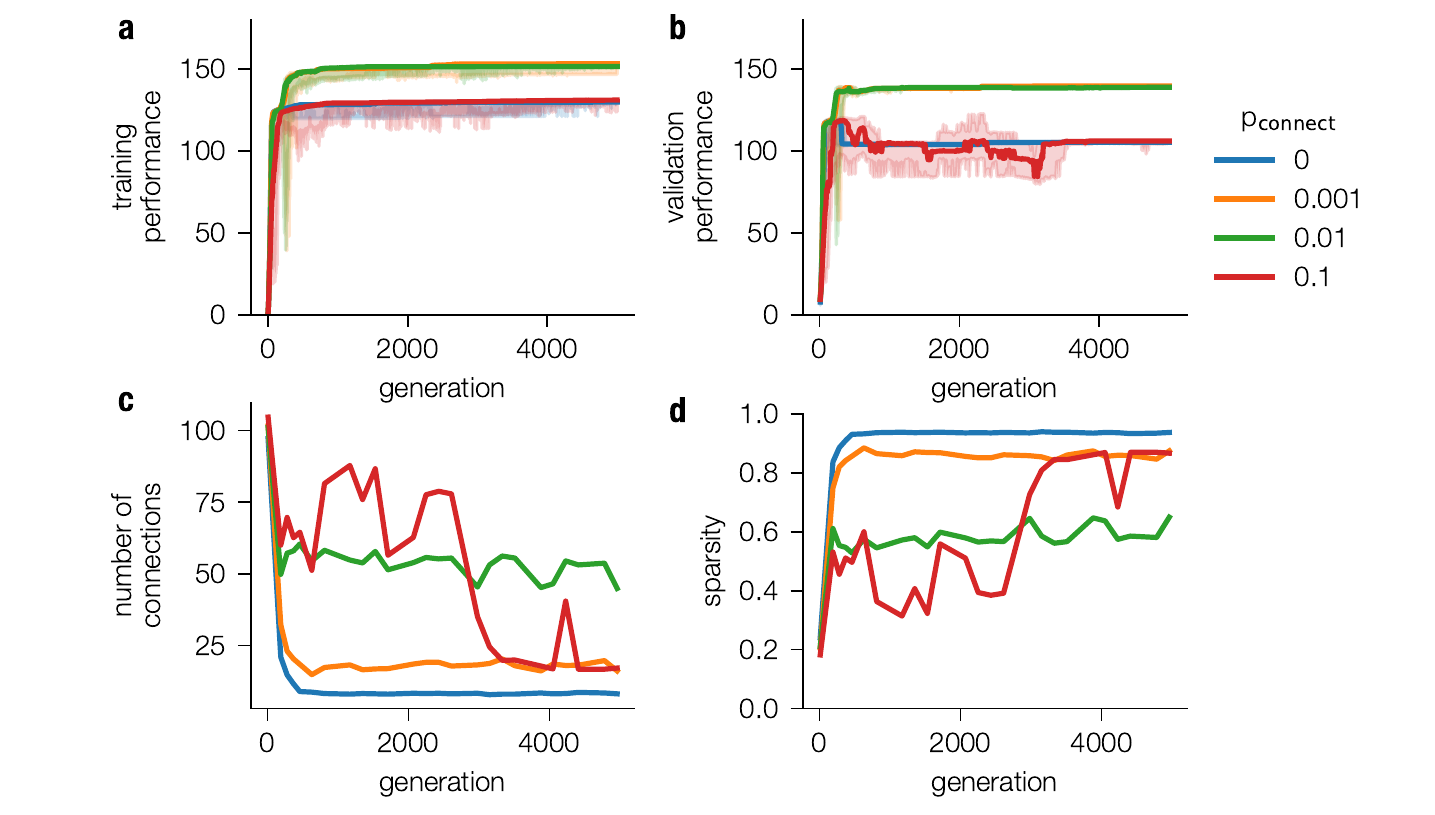}%
\caption{\hl{\textbf{Training/test performance as a function connection probability $p_\mathrm{connect}$ and disconnection probability $p_\mathrm{disconnect}$ ratio}}\newline
\hl{The training (a) and test (b) performance of the agents as a function of the generation for different values of $p_\mathrm{connect}$ (0, 0.001, 0.01, 0.1) and a fixed $p_\mathrm{disconnect} = 0.01$. Best performance is achieved when $p_\mathrm{connect}\leq p_\mathrm{disconnect}$ and both are in the same range. (c, d), The sparsity expressed as the number of connections (c) and as a relative count of realized connections (d). A connection is counted active, if the fitness is not decreased when the connection is disabled. Thus, the sparsity the networks is reduced by deleting sub-networks, which are not used (analogously to Supplements {\ref{fig:calc_control_to_sparse}}).}}
\label{fig:plotreconnect}
\end{figure}

\hl{Furthermore, we were able to show that training works best, in terms of validation and training performance, when the probability for removing existing connections $p_\mathrm{disconnect}$ is in the same range of the probability of recreating removed connections $p_\mathrm{connect}$ (Fig. {\ref{fig:plotreconnect}}). Thus, even when $p_\mathrm{disconnect}= p_\mathrm{connect}$ generalization performance of the networks is increased. Sparsity of the networks can additionally be increased by deleting all unused connections, which do not lead to any changes in performance (combination of pruning and deletion after training, c.f.~Fig {\ref{fig:plotreconnect}}). This, procedure can be sophisticated as the random severance of connections does not prevent the development of unconnected subnetworks, which should be finally deleted to remove pseudo-complexity of the neural networks. This could be done by simply thresholding the connection weights (c.f.~Supplements Fig. {\ref{fig:calc_control_to_sparse}}). However, gradually thresholding the network connections does not help to significantly increase the sparsity of fully connected networks (c.f.~control condition in Fig. {\ref{fig:calc_control_to_sparse}}) as these networks were not forced to distribute the information processing over a smaller amount of neurons by the random pruning procedure.}

\begin{figure}[htbp]
\centering
\makebox[\textwidth][c]{\includegraphics{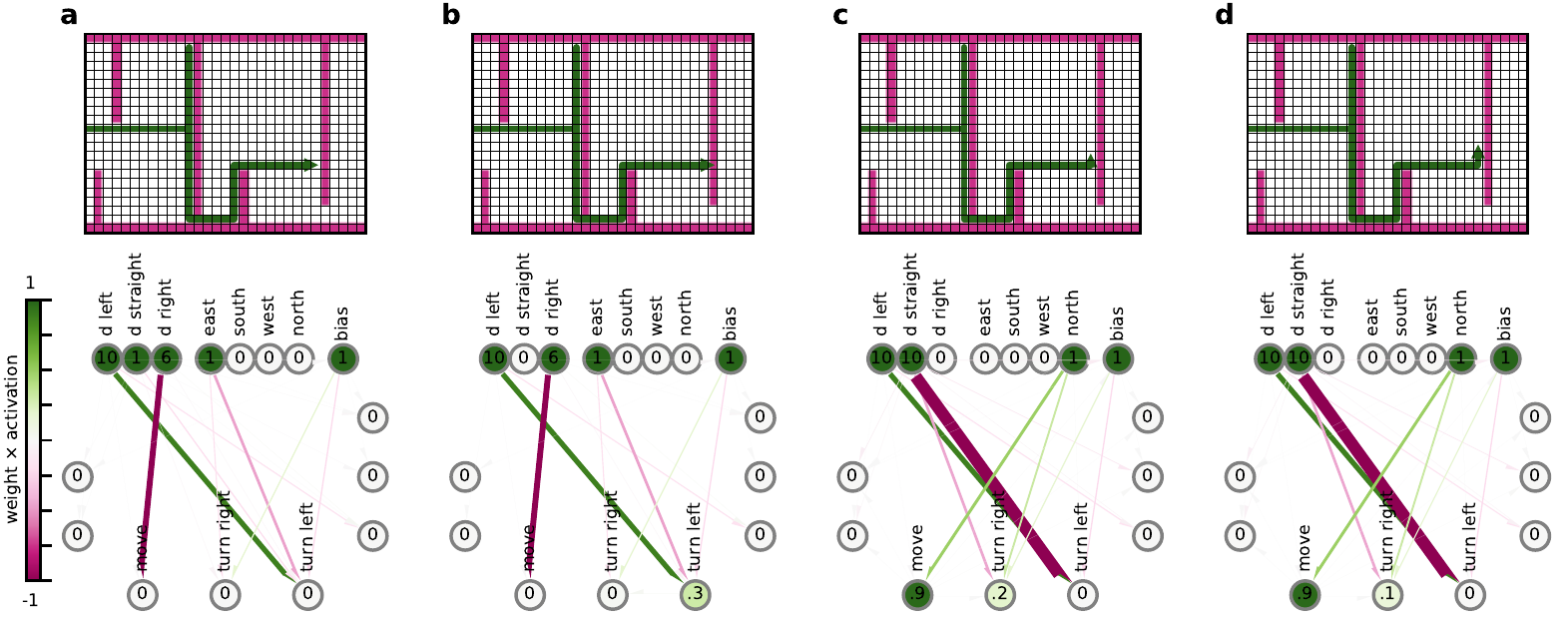}}%
\caption{\textbf{State of the maze and configuration of the network during a turn.} The position and orientation in the maze are denoted by the green triangle. The past trajectory by the green line. Below, the current state of the network is visualised. The values correspond to the current activation of each node and the colored connections to the weight times the activation \hl{(note that the weights are static)} of the target node (pink negative, green positive). \textbf{a}, Network one step before turn. \textbf{b}, Network encounters a wall and turns. \textbf{c}, Network has just turned and is now heading north. \textbf{d}, Network continues walking straight in the new direction.}
\label{fig:Turn}
\end{figure}

\section*{Conclusion}

In this study we showed that evolutionary pruning of artificial neural networks, evolutionary trained to solve a simple maze task, leads to sparser networks with better generalization properties compared to dense networks trained without pruning. The evolutionary pruning is realized via two mutation mechanisms: First the networks can set a threshold defining the upper limit of the absolute weight value being virtually set to zero. Secondly, some connections are removed (weight set to 0) \hl{and also restored} at random with given probabilities \hl{$p_\mathrm{connect}$, $p_\mathrm{disconnect}$}. The random removal of existing connections leads to more robust networks with better generalization abilities. \hl{This effect works best when $p_\mathrm{connect}\leq p_\mathrm{disconnect}$ and both probabilities are in the same range}.
In conclusion, evolutionary pruning by random severance of connections can be used as additional mechanism to improve the evolutionary training of neural networks. 

\section*{Discussion}
\subsection*{Limitations}
The maze task is still not complex enough to trigger the development of more complex recurrent neural networks which include abilities such as memory. As the agents in the maze are always provided with a ``compass'' meaning always know in which direction they are pointing, the task can be solved with a Markov like decision process, as the information of one time step is enough to decide the next action. Thus, after 5000 epochs the best performances are achieved by simple feed forward networks (cf. Fig. \ref{fig:Turn}). Nevertheless, it has to be considered that these networks were not forced to develop feed forward architectures but are a result of the Markov properties of the task they were trained on. The simulation shows that simple feed forward networks are able to navigate through a maze using only 16 sparsely connected neurons.

\subsection*{Future Research Directions}
\hl{Thus, the here described networks} demonstrate that only a small number of neurons is needed to navigate through environment and shows that for example \textit{C. elegans} with its 302 neurons \cite{J.G.WhiteE.SouthgateJ.N.Thomson1986} should be indeed able to perform complex tasks. It has already been shown that \textit{C. elegans} shows thermo- and electrotaxis \cite{Gabel2007} which are not simply a biased random walk in contrast to its' chemotaxis (for certain temperatures thermotaxis also changes to a random walk) \cite{Pierce-Shimomura1999, Ryu2002}. \textit{C. elegans} shows a direct movement along the electric gradient \cite{Gabel2007}. Thus, the amphid sensory neurons of \textit{C. elegans} could be seen as and equivalent to the "compass" neurons in our simulation. 

The here described study does not show a biologically inspired neuron configuration, however, demonstrates that 16 neurons are enough to perform relatively simple navigation tasks and gives a hint that the development of sparsity has evolutionary advantages. 
These findings contrast with current developments in the field of artificial intelligence where the size of the networks due to higher calculation power are scaled up \cite{Xu2018}.

Furthermore, the maze task could be extended so that the task is not a simple Markov process, which means that the next decision cannot be made by simply analyzing the current position in the maze \cite{Wierstra2004, Littman2012}. This increase of complexity can be achieved e.g. by removing the compass neurons. Consequently, a neural network which is able to achieve similar performance than networks with compass neurons need to develop some ``memory'' features. The networks would have to remember which movements they recently executed and they would have to dynamically recall this information. An even more demanding task would be to force the agents to go directly back to the starting point after they passed the maze. This task requires path integration \cite{wehner1986path, etienne2004path}, which in turn requires the ability to flexibly navigate in physical space like insects \cite{wehner1986path, muller1988path, muller1994hidden, andel2004path}, and at least in mammals episodic memory \cite{etienne1996path, seguinot1998path, mcnaughton2006path}. It has been demonstrated that these abilities can only be achieved within highly recurrent networks, as they can be found in the hippocampus \cite{etienne2004path}. Recently it has been shown that the network architecture of the hippocampus is not limited to spatial navigation, but seems to be domain-general \cite{aronov2017mapping, killian2018grid, nau2018hexadirectional} and even allows navigation in abstract high-dimensional cognitive feature spaces \cite{eichenbaum2015hippocampus, constantinescu2016organizing, garvert2017map, bellmund2018navigating, theves2019hippocampus}. Future work will have to investigate, whether networks with the above mentioned abilities can also be found by evolutionary algorithms. 

\hl{Further potential research directions could be to optimize network architectures using evolutionary algorithms to find efficient neural networks, which can be trained supervisely or via reinforcement learning. The counterpart to this approach would be the analysis of evolutionary trained sparse neural networks, and to search for functional units such as motifs \mbox{\cite{krauss2019analysis}}, weight statistics \mbox{\cite{krauss2019weight}}, or to analyze complex dynamics like 'recurrence resonance' effects \mbox{\cite{krauss2019recurrence}}, which is up to now often done in untrained networks, that perform no real information processing.}

\hl{These two strands, would be in line with the philosophy that artificial and biological intelligence are "two sides of the same coin" \mbox{\cite{schilling2018deep, Kriegeskorte2018}}, and that the understanding of brain mechanisms on the one hand and the development of artificial neural network algorithms on the other hand are an iterative process, stimulating each other.}

\section*{Acknowledgements}
This work was supported by the German Research Foundation (DFG, grant KR5148/2-1 to PK), and the Emergent Talents Initiative (ETI) of the University Erlangen-Nuremberg (grant 2019/2-Phil-01 to PK).
                
\section*{Author Contributions}
RG and AS designed the study. RG and AS performed the simulations. RG, AS, AE and PK discussed the results. RG, AS and PK wrote the manuscript. 

\bibliographystyle{apa}
\bibliography{references}

\begin{thebibliography}{}

\bibitem[\protect\astroncite{Alexandre et~al.}{2009}]{Alexandre2009}
Alexandre, L.~A., Embrechts, M.~J., and Linton, J. (2009).
\newblock {Benchmarking reservoir computing on time-independent classification
  tasks}.
\newblock {\em Proceedings of the International Joint Conference on Neural
  Networks}, pages 89--93.

\bibitem[\protect\astroncite{Amaral et~al.}{2000}]{Amaral2000}
Amaral, L.~A., Scala, A., Barth{\'{e}}l{\'{e}}my, M., and Stanley, H.~E.
  (2000).
\newblock {Classes of small-world networks}.
\newblock {\em Proceedings of the National Academy of Sciences of the United
  States of America}, 97(21):11149--11152.

\bibitem[\protect\astroncite{Andel and Wehner}{2004}]{andel2004path}
Andel, D. and Wehner, R. (2004).
\newblock Path integration in desert ants, cataglyphis: how to make a homing
  ant run away from home.
\newblock {\em Proceedings of the Royal Society of London. Series B: Biological
  Sciences}, 271(1547):1485--1489.

\bibitem[\protect\astroncite{Antonelo and Schrauwen}{2012}]{Antonelo2012}
Antonelo, E. and Schrauwen, B. (2012).
\newblock {Learning Slow Features with Reservoir Networks for
  Biologically-inspired Robot Localization}.
\newblock {\em Neural Networks}, 25:178--190.

\bibitem[\protect\astroncite{Antonelo et~al.}{2007}]{Antonelo2007}
Antonelo, E.~A., Schrauwen, B., and Stroobandt, D. (2007).
\newblock {Event detection and localization for small mobile robots using
  reservoir computing}.
\newblock In {\em Conference on Artificial Neural Networks}, pages 660--669.

\bibitem[\protect\astroncite{Anwar et~al.}{2017}]{Anwar2017}
Anwar, S., Hwang, K., and Sung, W. (2017).
\newblock {Structured pruning of deep convolutional neural networks}.
\newblock {\em ACM Journal on Emerging Technologies in Computing Systems},
  13(3):1--18.

\bibitem[\protect\astroncite{Aronov et~al.}{2017}]{aronov2017mapping}
Aronov, D., Nevers, R., and Tank, D.~W. (2017).
\newblock Mapping of a non-spatial dimension by the hippocampal--entorhinal
  circuit.
\newblock {\em Nature}, 543(7647):719.

\bibitem[\protect\astroncite{Babadi and
  Sompolinsky}{2014}]{babadi2014sparseness}
Babadi, B. and Sompolinsky, H. (2014).
\newblock Sparseness and expansion in sensory representations.
\newblock {\em Neuron}, 83(5):1213--1226.

\bibitem[\protect\astroncite{Barwich}{2019}]{barwich2019value}
Barwich, A.-S. (2019).
\newblock The value of failure in science: The story of grandmother cells in
  neuroscience.
\newblock {\em Frontiers in neuroscience}, 13:1121.

\bibitem[\protect\astroncite{Bassett and Bullmore}{2006}]{Bassett2006}
Bassett, D.~S. and Bullmore, E. (2006).
\newblock {Small-world brain networks}.
\newblock {\em Neuroscientist}, 12(6):512--523.

\bibitem[\protect\astroncite{Bassett et~al.}{2006}]{Bullmore2006}
Bassett, D.~S., Meyer-Lindenberg, A., Achard, S., Duke, T., and Bullmore, E.
  (2006).
\newblock {Adaptive reconfiguration of fractal small-world human brain
  functional networks}.
\newblock {\em Proceedings of the National Academy of Sciences},
  103(51):19518--19523.

\bibitem[\protect\astroncite{Bellmund et~al.}{2018}]{bellmund2018navigating}
Bellmund, J.~L., G{\"a}rdenfors, P., Moser, E.~I., and Doeller, C.~F. (2018).
\newblock Navigating cognition: Spatial codes for human thinking.
\newblock {\em Science}, 362(6415):eaat6766.

\bibitem[\protect\astroncite{Bertschinger and
  Natschl{\"{a}}ger}{2004}]{Bertschinger2004}
Bertschinger, N. and Natschl{\"{a}}ger, T. (2004).
\newblock {Real-time computation at the edge of chaos in recurrent neural
  networks}.
\newblock {\em Neural Computation}, 16(7):1413--1436.

\bibitem[\protect\astroncite{Boksa}{2012}]{boksa2012abnormal}
Boksa, P. (2012).
\newblock Abnormal synaptic pruning in schizophrenia: Urban myth or reality?
\newblock {\em Journal of psychiatry \& neuroscience: JPN}, 37(2):75.

\bibitem[\protect\astroncite{Chalfie}{1984}]{chalfie1984neuronal}
Chalfie, M. (1984).
\newblock Neuronal development in caenorhabditis elegans.
\newblock {\em Trends in NeuroSciences}, 7(6):197--202.

\bibitem[\protect\astroncite{Constantinescu
  et~al.}{2016}]{constantinescu2016organizing}
Constantinescu, A.~O., O’Reilly, J.~X., and Behrens, T.~E. (2016).
\newblock Organizing conceptual knowledge in humans with a gridlike code.
\newblock {\em Science}, 352(6292):1464--1468.

\bibitem[\protect\astroncite{Crochet et~al.}{2011}]{crochet2011synaptic}
Crochet, S., Poulet, J.~F., Kremer, Y., and Petersen, C.~C. (2011).
\newblock Synaptic mechanisms underlying sparse coding of active touch.
\newblock {\em Neuron}, 69(6):1160--1175.

\bibitem[\protect\astroncite{Dasgupta et~al.}{2017}]{dasgupta2017neural}
Dasgupta, S., Stevens, C.~F., and Navlakha, S. (2017).
\newblock A neural algorithm for a fundamental computing problem.
\newblock {\em Science}, 358(6364):793--796.

\bibitem[\protect\astroncite{Eichenbaum}{2015}]{eichenbaum2015hippocampus}
Eichenbaum, H. (2015).
\newblock The hippocampus as a cognitive map… of social space.
\newblock {\em Neuron}, 87(1):9--11.

\bibitem[\protect\astroncite{Etienne and Jeffery}{2004}]{etienne2004path}
Etienne, A.~S. and Jeffery, K.~J. (2004).
\newblock Path integration in mammals.
\newblock {\em Hippocampus}, 14(2):180--192.

\bibitem[\protect\astroncite{Etienne et~al.}{1996}]{etienne1996path}
Etienne, A.~S., Maurer, R., and S{\'e}guinot, V. (1996).
\newblock Path integration in mammals and its interaction with visual
  landmarks.
\newblock {\em Journal of experimental Biology}, 199(1):201--209.

\bibitem[\protect\astroncite{Fekiac et~al.}{2011}]{Fekiac2011}
Fekiac, J., Zelinka, I., and Burguillo, J.~C. (2011).
\newblock {A review of methods for encoding neural network topologies in
  evolutionary computation}.
\newblock {\em Proceedings - 25th European Conference on Modelling and
  Simulation, ECMS 2011}, pages 410--416.

\bibitem[\protect\astroncite{Gabel et~al.}{2007}]{Gabel2007}
Gabel, C.~V., Gabel, H., Pavlichin, D., Kao, A., Clark, D.~A., and Samuel,
  A.~D. (2007).
\newblock {Neural Circuits Mediate Electrosensory Behavior in Caenorhabditis
  elegans}.
\newblock {\em Journal of Neuroscience}, 27(28):7586--7596.

\bibitem[\protect\astroncite{Garvert et~al.}{2017}]{garvert2017map}
Garvert, M.~M., Dolan, R.~J., and Behrens, T.~E. (2017).
\newblock A map of abstract relational knowledge in the human
  hippocampal--entorhinal cortex.
\newblock {\em Elife}, 6:e17086.

\bibitem[\protect\astroncite{Gerum}{2019}]{Gerum2019}
Gerum, R. (2019).
\newblock {pylustrator: Code generation for reproducible figures for
  publication}.
\newblock {\em arXiv}, 1910.00279.

\bibitem[\protect\astroncite{Hagmann et~al.}{2008}]{Hagmann2008}
Hagmann, P., Cammoun, L., Gigandet, X., Meuli, R., Honey, C.~J., {Van Wedeen},
  J., and Sporns, O. (2008).
\newblock {Mapping the structural core of human cerebral cortex}.
\newblock {\em PLoS Biology}, 6(7):1479--1493.

\bibitem[\protect\astroncite{Han et~al.}{2015}]{Han2015}
Han, S., Mao, H., and Dally, W.~J. (2015).
\newblock {Deep Compression: Compressing Deep Neural Networks with Pruning,
  Trained Quantization and Huffman Coding}.
\newblock {\em arXiv:1510.00149}, pages 1--14.

\bibitem[\protect\astroncite{Herculano-Houzel}{2009}]{herculano2009human}
Herculano-Houzel, S. (2009).
\newblock The human brain in numbers: a linearly scaled-up primate brain.
\newblock {\em Frontiers in human neuroscience}, 3:31.

\bibitem[\protect\astroncite{Hochreiter}{1998}]{Hochreiter1998}
Hochreiter, S. (1998).
\newblock {The Vanishing Gradient Problem During Learning Recurrent Neural Nets
  and Problem Solutions}.
\newblock {\em International Journal of Uncertainty, Fuzziness and
  Knowledge-Based Systems}, 6(2):107--116.

\bibitem[\protect\astroncite{Hong et~al.}{2016}]{hong2016new}
Hong, S., Dissing-Olesen, L., and Stevens, B. (2016).
\newblock New insights on the role of microglia in synaptic pruning in health
  and disease.
\newblock {\em Current opinion in neurobiology}, 36:128--134.

\bibitem[\protect\astroncite{Hunter}{2007}]{Hunter2007a}
Hunter, J.~D. (2007).
\newblock {Matplotlib: A 2D graphics environment}.
\newblock {\em Computing in Science and Engineering}, 9:90--95.

\bibitem[\protect\astroncite{Jarrell et~al.}{2012}]{jarrell2012connectome}
Jarrell, T.~A., Wang, Y., Bloniarz, A.~E., Brittin, C.~A., Xu, M., Thomson,
  J.~N., Albertson, D.~G., Hall, D.~H., and Emmons, S.~W. (2012).
\newblock The connectome of a decision-making neural network.
\newblock {\em Science}, 337(6093):437--444.

\bibitem[\protect\astroncite{{J.G.White, E.Southgate,
  J.N.Thomson}}{1986}]{J.G.WhiteE.SouthgateJ.N.Thomson1986}
{J.G.White, E.Southgate, J.N.Thomson}, S. (1986).
\newblock {The Structure of the Nervous System of the Nematode Caenorhabditis
  elegans Author ( s ): J . G . White , E . Southgate , J . N . Thomson , S .
  Brenner Source : Philosophical Transactions of the Royal Society of London .
  Series B , Biological Published by}.
\newblock {\em Philosophical Transactions of the Royal Society of London},
  314(1165):1--340.

\bibitem[\protect\astroncite{Jiao et~al.}{2018}]{jiao2018sparse}
Jiao, Y., Zhang, Y., Chen, X., Yin, E., Jin, J., Wang, X., and Cichocki, A.
  (2018).
\newblock Sparse group representation model for motor imagery eeg
  classification.
\newblock {\em IEEE journal of biomedical and health informatics},
  23(2):631--641.

\bibitem[\protect\astroncite{Jin et~al.}{2018}]{jin2018eeg}
Jin, Z., Zhou, G., Gao, D., and Zhang, Y. (2018).
\newblock Eeg classification using sparse bayesian extreme learning machine for
  brain--computer interface.
\newblock {\em Neural Computing and Applications}, pages 1--9.

\bibitem[\protect\astroncite{Jonas and Kording}{2017}]{Jonas2017}
Jonas, E. and Kording, K.~P. (2017).
\newblock Could a neuroscientist understand a microprocessor?
\newblock {\em PLoS computational biology}, 13(1):e1005268.

\bibitem[\protect\astroncite{Kafashan et~al.}{2016}]{kafashan2016relating}
Kafashan, M., Nandi, A., and Ching, S. (2016).
\newblock Relating observability and compressed sensing of time-varying signals
  in recurrent linear networks.
\newblock {\em Neural Networks}, 83:11--20.

\bibitem[\protect\astroncite{Kerr et~al.}{2005}]{Kerr2005}
Kerr, J.~N., Greenberg, D., and Helmchen, F. (2005).
\newblock {Imaging input and output of neocortical networks in vivo}.
\newblock {\em Proc Natl Acad Sci USA}, 102(39):14063--14068.

\bibitem[\protect\astroncite{Killian and Buffalo}{2018}]{killian2018grid}
Killian, N.~J. and Buffalo, E.~A. (2018).
\newblock Grid cells map the visual world.
\newblock {\em Nature neuroscience}, 21(2):161.

\bibitem[\protect\astroncite{Kolb and Gibb}{2011}]{kolb2011brain}
Kolb, B. and Gibb, R. (2011).
\newblock Brain plasticity and behaviour in the developing brain.
\newblock {\em Journal of the Canadian Academy of Child and Adolescent
  Psychiatry}, 20(4):265.

\bibitem[\protect\astroncite{Krauss et~al.}{2019a}]{krauss2019recurrence}
Krauss, P., Prebeck, K., Schilling, A., and Metzner, C. (2019a).
\newblock Recurrence resonance” in three-neuron motifs.
\newblock {\em Frontiers in computational neuroscience}, 13.

\bibitem[\protect\astroncite{Krauss et~al.}{2019b}]{Krauss2019}
Krauss, P., Schuster, M., Dietrich, V., Schilling, A., Schulze, H., and
  Metzner, C. (2019b).
\newblock {Weight statistics controls dynamics in recurrent neural networks}.
\newblock {\em PLoS ONE}, 14(4):1--13.

\bibitem[\protect\astroncite{Krauss et~al.}{2019c}]{krauss2019weight}
Krauss, P., Schuster, M., Dietrich, V., Schilling, A., Schulze, H., and
  Metzner, C. (2019c).
\newblock Weight statistics controls dynamics in recurrent neural networks.
\newblock {\em PloS one}, 14(4).

\bibitem[\protect\astroncite{Krauss et~al.}{2019d}]{krauss2019analysis}
Krauss, P., Zankl, A., Schilling, A., Schulze, H., and Metzner, C. (2019d).
\newblock Analysis of structure and dynamics in three-neuron motifs.
\newblock {\em Frontiers in computational neuroscience}, 13:5.

\bibitem[\protect\astroncite{Kriegeskorte and Douglas}{2018}]{Kriegeskorte2018}
Kriegeskorte, N. and Douglas, P.~K. (2018).
\newblock {Cognitive computational neuroscience}.
\newblock {\em Nature Neuroscience}, 21(9):1148--1160.

\bibitem[\protect\astroncite{Latora and Marchiori}{2001}]{latora2001efficient}
Latora, V. and Marchiori, M. (2001).
\newblock Efficient behavior of small-world networks.
\newblock {\em Physical review letters}, 87(19):198701.

\bibitem[\protect\astroncite{Littman}{2012}]{Littman2012}
Littman, M. (2012).
\newblock {Inducing Partially Observable Markov Decision Processes.}
\newblock {\em 11th International Conference on Grammatical Inference},
  21:145--148.

\bibitem[\protect\astroncite{Low and Cheng}{2006}]{low2006axon}
Low, L.~K. and Cheng, H.-J. (2006).
\newblock Axon pruning: an essential step underlying the developmental
  plasticity of neuronal connections.
\newblock {\em Philosophical Transactions of the Royal Society B: Biological
  Sciences}, 361(1473):1531--1544.

\bibitem[\protect\astroncite{Luko{\v{s}}evi{\v{c}}ius
  et~al.}{2012}]{Lukosevicius2012}
Luko{\v{s}}evi{\v{c}}ius, M., Jaeger, H., and Schrauwen, B. (2012).
\newblock {Reservoir Computing Trends}.
\newblock {\em KI - K{\"{u}}nstliche Intelligenz}, 26(4):365--371.

\bibitem[\protect\astroncite{McNaughton et~al.}{2006}]{mcnaughton2006path}
McNaughton, B.~L., Battaglia, F.~P., Jensen, O., Moser, E.~I., and Moser, M.-B.
  (2006).
\newblock Path integration and the neural basis of the'cognitive map'.
\newblock {\em Nature Reviews Neuroscience}, 7(8):663.

\bibitem[\protect\astroncite{Mocanu et~al.}{2018}]{mocanu2018scalable}
Mocanu, D.~C., Mocanu, E., Stone, P., Nguyen, P.~H., Gibescu, M., and Liotta,
  A. (2018).
\newblock Scalable training of artificial neural networks with adaptive sparse
  connectivity inspired by network science.
\newblock {\em Nature communications}, 9(1):2383.

\bibitem[\protect\astroncite{M{\"u}ller and Wehner}{1988}]{muller1988path}
M{\"u}ller, M. and Wehner, R. (1988).
\newblock Path integration in desert ants, cataglyphis fortis.
\newblock {\em Proceedings of the National Academy of Sciences},
  85(14):5287--5290.

\bibitem[\protect\astroncite{M{\"u}ller and Wehner}{1994}]{muller1994hidden}
M{\"u}ller, M. and Wehner, R. (1994).
\newblock The hidden spiral: systematic search and path integration in desert
  ants, cataglyphis fortis.
\newblock {\em Journal of Comparative Physiology A}, 175(5):525--530.

\bibitem[\protect\astroncite{Nau et~al.}{2018}]{nau2018hexadirectional}
Nau, M., Schr{\"o}der, T.~N., Bellmund, J.~L., and Doeller, C.~F. (2018).
\newblock Hexadirectional coding of visual space in human entorhinal cortex.
\newblock {\em Nature neuroscience}, 21(2):188.

\bibitem[\protect\astroncite{Oh et~al.}{2014}]{Oh2014}
Oh, S.~W., Harris, J.~A., Ng, L., Winslow, B., Cain, N., Mihalas, S., Wang, Q.,
  Lau, C., Kuan, L., Henry, A.~M., Mortrud, M.~T., Ouellette, B., Nguyen,
  T.~N., Sorensen, S.~A., Slaughterbeck, C.~R., Wakeman, W., Li, Y., Feng, D.,
  Ho, A., Nicholas, E., Hirokawa, K.~E., Bohn, P., Joines, K.~M., Peng, H.,
  Hawrylycz, M.~J., Phillips, J.~W., Hohmann, J.~G., Wohnoutka, P., Gerfen,
  C.~R., Koch, C., Bernard, A., Dang, C., Jones, A.~R., and Zeng, H. (2014).
\newblock {A mesoscale connectome of the mouse brain}.
\newblock {\em Nature}, 508(7495):207--214.

\bibitem[\protect\astroncite{Olshausen and Field}{2004}]{olshausen2004sparse}
Olshausen, B.~A. and Field, D.~J. (2004).
\newblock Sparse coding of sensory inputs.
\newblock {\em Current opinion in neurobiology}, 14(4):481--487.

\bibitem[\protect\astroncite{Oren-Suissa et~al.}{2016}]{oren2016sex}
Oren-Suissa, M., Bayer, E.~A., and Hobert, O. (2016).
\newblock Sex-specific pruning of neuronal synapses in caenorhabditis elegans.
\newblock {\em Nature}, 533(7602):206.

\bibitem[\protect\astroncite{Paolicelli et~al.}{2011}]{paolicelli2011synaptic}
Paolicelli, R.~C., Bolasco, G., Pagani, F., Maggi, L., Scianni, M., Panzanelli,
  P., Giustetto, M., Ferreira, T.~A., Guiducci, E., Dumas, L., et~al. (2011).
\newblock Synaptic pruning by microglia is necessary for normal brain
  development.
\newblock {\em science}, 333(6048):1456--1458.

\bibitem[\protect\astroncite{Pascanu et~al.}{2012}]{Pascanu2012}
Pascanu, R., Mikolov, T., and Bengio, Y. (2012).
\newblock {Understanding the exploding gradient problem}.
\newblock {\em arXiv:1211.5063v1}.

\bibitem[\protect\astroncite{Pehlevan and
  Sompolinsky}{2014}]{pehlevan2014selectivity}
Pehlevan, C. and Sompolinsky, H. (2014).
\newblock Selectivity and sparseness in randomly connected balanced networks.
\newblock {\em PLoS One}, 9(2).

\bibitem[\protect\astroncite{Perin et~al.}{2011}]{Perin2011}
Perin, R., Berger, T.~K., and Markram, H. (2011).
\newblock {A synaptic organizing principle for cortical neuronal groups}.
\newblock {\em Proceedings of the National Academy of Sciences},
  108(13):5419--5424.

\bibitem[\protect\astroncite{Pierce-Shimomura
  et~al.}{1999}]{Pierce-Shimomura1999}
Pierce-Shimomura, J.~T., Morse, T.~M., and Lockery, S.~R. (1999).
\newblock {The fundamental role of pirouettes in Caenorhabditis elegans
  chemotaxis}.
\newblock {\em Journal of Neuroscience}, 19(21):9557--9569.

\bibitem[\protect\astroncite{Quiroga et~al.}{2008}]{quiroga2008sparse}
Quiroga, R.~Q., Kreiman, G., Koch, C., and Fried, I. (2008).
\newblock Sparse but not ‘grandmother-cell’coding in the medial temporal
  lobe.
\newblock {\em Trends in cognitive sciences}, 12(3):87--91.

\bibitem[\protect\astroncite{Rose}{1996}]{rose1996some}
Rose, D. (1996).
\newblock Some reflections on (or by?) grandmother cells.

\bibitem[\protect\astroncite{Ryu and Samuel}{2002}]{Ryu2002}
Ryu, W.~S. and Samuel, A. D.~T. (2002).
\newblock {Thermotaxis in Caenorhabditis elegans Analyzed by Measuring
  Responses to Defined Thermal Stimuli}.
\newblock {\em Journal of Neuroscience}, 22(13):1--7.

\bibitem[\protect\astroncite{Sanchez et~al.}{2001}]{sanchez2001solving}
Sanchez, E., P{\'e}rez-Uribe, A., and Mesot, B. (2001).
\newblock Solving partially observable problems by evolution and learning of
  finite state machines.
\newblock In {\em International Conference on Evolvable Systems}, pages
  267--278. Springer.

\bibitem[\protect\astroncite{Schilling et~al.}{2018}]{schilling2018deep}
Schilling, A., Metzner, C., Rietsch, J., Gerum, R., Schulze, H., and Krauss, P.
  (2018).
\newblock How deep is deep enough?--quantifying class separability in the
  hidden layers of deep neural networks.
\newblock {\em arXiv preprint arXiv:1811.01753}.

\bibitem[\protect\astroncite{Schmidhuber and
  Hochreiter}{1997}]{Schmidhuber1997}
Schmidhuber, J. and Hochreiter, S. (1997).
\newblock {Long short-term memory}.
\newblock {\em Neural Computation}, 9(8):1735--1780.

\bibitem[\protect\astroncite{Schrauwen et~al.}{2007}]{Schrauwen2007}
Schrauwen, B., Verstraeten, D., and {Van Campenhout}, J. (2007).
\newblock {An overview of reservoir computing: theory, applications and
  implementations}.
\newblock {\em Proceedings of the 15th European Symposium on Artificial Neural
  Networks}, pages 471--82.

\bibitem[\protect\astroncite{S{\'e}guinot et~al.}{1998}]{seguinot1998path}
S{\'e}guinot, V., Cattet, J., and Benhamou, S. (1998).
\newblock Path integration in dogs.
\newblock {\em Animal behaviour}, 55(4):787--797.

\bibitem[\protect\astroncite{Song et~al.}{2005}]{Song2005}
Song, S., Sj{\"{o}}str{\"{o}}m, P.~J., Reigl, M., Nelson, S., and Chklovskii,
  D.~B. (2005).
\newblock {Highly nonrandom features of synaptic connectivity in local cortical
  circuits}.
\newblock {\em PLoS Biology}, 3(3):0507--0519.

\bibitem[\protect\astroncite{Sporns et~al.}{2005}]{Sporns2005}
Sporns, O., Tononi, G., and K{\"{o}}tter, R. (2005).
\newblock {The human connectome: A structural description of the human brain}.
\newblock {\em PLoS Computational Biology}, 1(4):0245--0251.

\bibitem[\protect\astroncite{Springenberg
  et~al.}{2016}]{springenberg2016bayesian}
Springenberg, J.~T., Klein, A., Falkner, S., and Hutter, F. (2016).
\newblock Bayesian optimization with robust bayesian neural networks.
\newblock In {\em Advances in neural information processing systems}, pages
  4134--4142.

\bibitem[\protect\astroncite{Theves et~al.}{2019}]{theves2019hippocampus}
Theves, S., Fernandez, G., and Doeller, C.~F. (2019).
\newblock The hippocampus encodes distances in multidimensional feature space.
\newblock {\em Current Biology}, 29(7):1226--1231.

\bibitem[\protect\astroncite{van~den Heuvel and Yeo}{2017}]{VandenHeuvel2017}
van~den Heuvel, M.~P. and Yeo, B.~T. (2017).
\newblock {A Spotlight on Bridging Microscale and Macroscale Human Brain
  Architecture}.
\newblock {\em Neuron}, 93(6):1248--1251.

\bibitem[\protect\astroncite{{Van Der Walt} et~al.}{2011}]{VanDerWalt2011b}
{Van Der Walt}, S., Colbert, S.~C., and Varoquaux, G. (2011).
\newblock {The NumPy array: A structure for efficient numerical computation}.
\newblock {\em Computing in Science and Engineering}, 13:22--30.

\bibitem[\protect\astroncite{Verstraeten et~al.}{2007}]{Verstraeten2007}
Verstraeten, D., Schrauwen, B., D'Haene, M., and Stroobandt, D. (2007).
\newblock {An experimental unification of reservoir computing methods}.
\newblock {\em Neural Networks}, 20(3):391--403.

\bibitem[\protect\astroncite{Watts and Strogatz}{1998}]{Watts1998}
Watts, D.~J. and Strogatz, S.~H. (1998).
\newblock {Strogatz - small world network Nature}.
\newblock {\em Nature}, 393:440--442.

\bibitem[\protect\astroncite{Wehner and Wehner}{1986}]{wehner1986path}
Wehner, R. and Wehner, S. (1986).
\newblock Path integration in desert ants. approaching a long-standing puzzle
  in insect navigation.
\newblock {\em Monitore Zoologico Italiano-Italian Journal of Zoology},
  20(3):309--331.

\bibitem[\protect\astroncite{Wen et~al.}{2016}]{Wen2016}
Wen, W., Wu, C., Wang, Y., Chen, Y., and Li, H. (2016).
\newblock Learning structured sparsity in deep neural networks.
\newblock In {\em Advances in neural information processing systems}, pages
  2074--2082.

\bibitem[\protect\astroncite{Wierstra and Wiering}{2004}]{Wierstra2004}
Wierstra, D. and Wiering, M. (2004).
\newblock {Utile distinction hidden Markov models}.
\newblock {\em Proceedings, Twenty-First International Conference on Machine
  Learning, ICML 2004}, pages 855--862.

\bibitem[\protect\astroncite{Xu et~al.}{2018}]{Xu2018}
Xu, X., Ding, Y., Hu, S.~X., Niemier, M., Cong, J., Hu, Y., and Shi, Y. (2018).
\newblock {Scaling for edge inference of deep neural networks}.
\newblock {\em Nature Electronics}, 1(4):216--222.

\bibitem[\protect\astroncite{Yeo and Gautier}{2004}]{yeo2004early}
Yeo, W. and Gautier, J. (2004).
\newblock Early neural cell death: dying to become neurons.
\newblock {\em Developmental biology}, 274(2):233--244.

\bibitem[\protect\astroncite{Young et~al.}{2015}]{young2015optimizing}
Young, S.~R., Rose, D.~C., Karnowski, T.~P., Lim, S.-H., and Patton, R.~M.
  (2015).
\newblock Optimizing deep learning hyper-parameters through an evolutionary
  algorithm.
\newblock In {\em Proceedings of the Workshop on Machine Learning in
  High-Performance Computing Environments}, pages 1--5.

\bibitem[\protect\astroncite{Zaslaver et~al.}{2015}]{zaslaver2015hierarchical}
Zaslaver, A., Liani, I., Shtangel, O., Ginzburg, S., Yee, L., and Sternberg,
  P.~W. (2015).
\newblock Hierarchical sparse coding in the sensory system of caenorhabditis
  elegans.
\newblock {\em Proceedings of the National Academy of Sciences},
  112(4):1185--1189.

\bibitem[\protect\astroncite{Zhang et~al.}{2011}]{zhang2011sparse}
Zhang, L., Yang, M., and Feng, X. (2011).
\newblock Sparse representation or collaborative representation: Which helps
  face recognition?
\newblock In {\em 2011 International conference on computer vision}, pages
  471--478. IEEE.

\end{thebibliography}

\newpage
\section*{Supplementary Material}
\renewcommand\thefigure{S\arabic{figure}}    
\setcounter{figure}{0}  

\begin{figure}[htbp]
\centering
\makebox[\textwidth][c]{\includegraphics[width=1.25\textwidth]{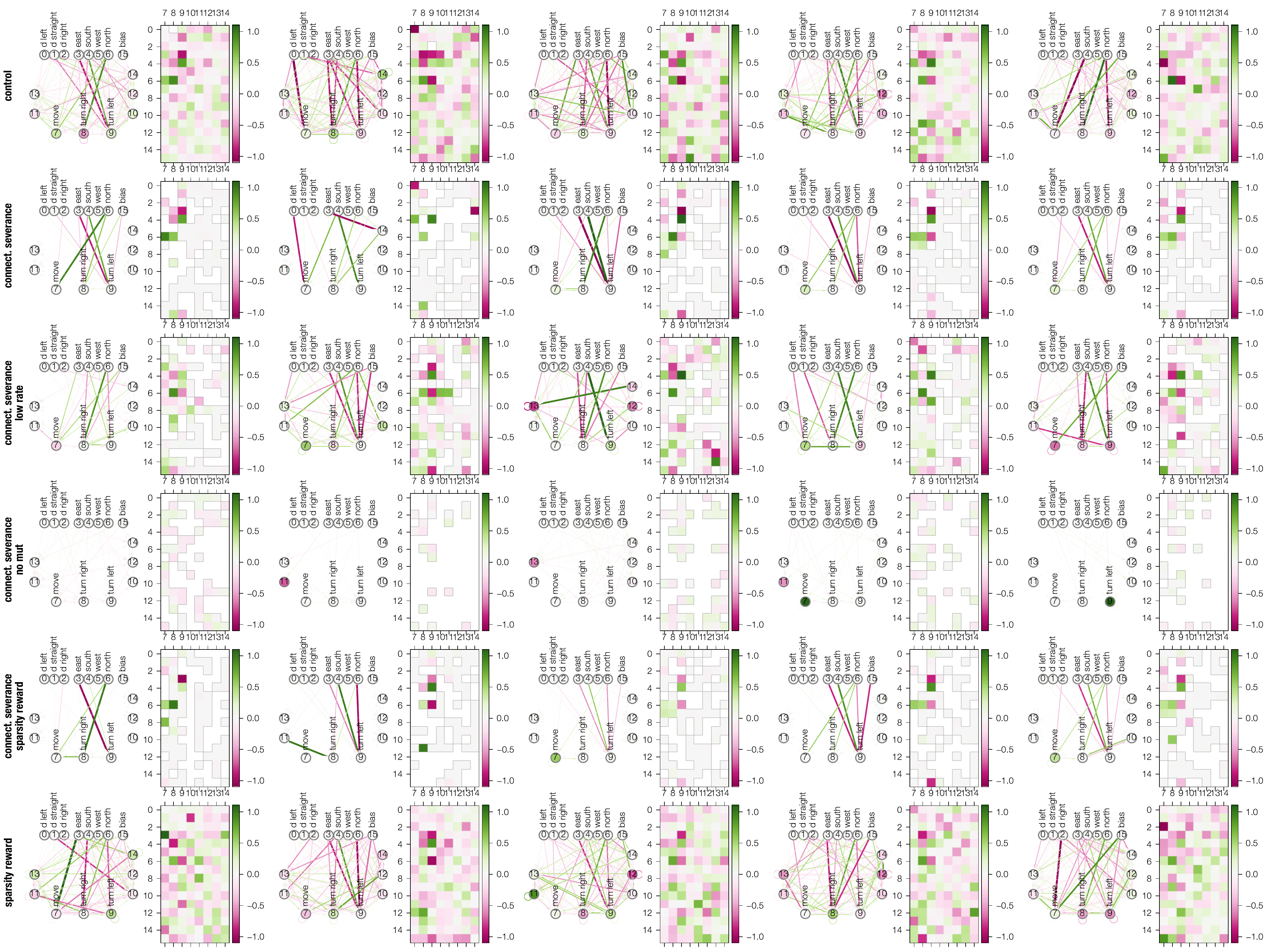}}%
\caption{\textbf{Best network connections for all experiments and seeds.} The network connections of the best network visualised as a connected graph and a connection matrix for each seed (column) and each experiment (row).}
\label{fig:AllConnectionMatrices}
\end{figure}

\begin{figure}[htbp]
\centering
\makebox[\textwidth][c]{\includegraphics[width=1.1\textwidth]{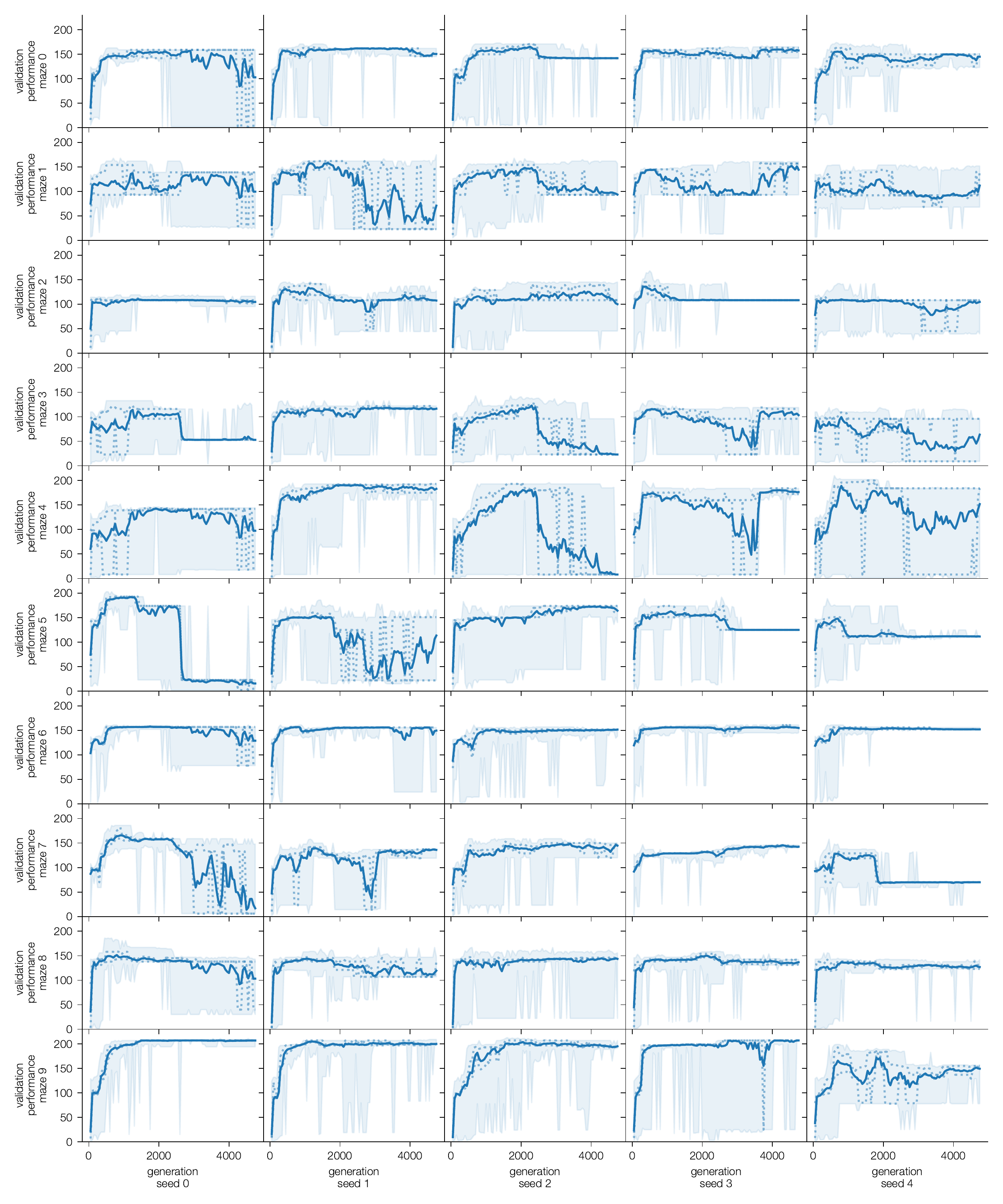}}%
\caption{\textbf{Validation performance in the "control" experiment.} The validation performance over generation for different seeds (columns) and different validation mazes (rows). Shaded area denotes the range from minimum to maximum, dotted lines indicate the 25\% and 75\% percentile, and the solid line denotes the mean.}
\label{fig:ValidationFitnessContol}
\end{figure}
\begin{figure}[htbp]
\centering
\makebox[\textwidth][c]{\includegraphics[width=1.1\textwidth]{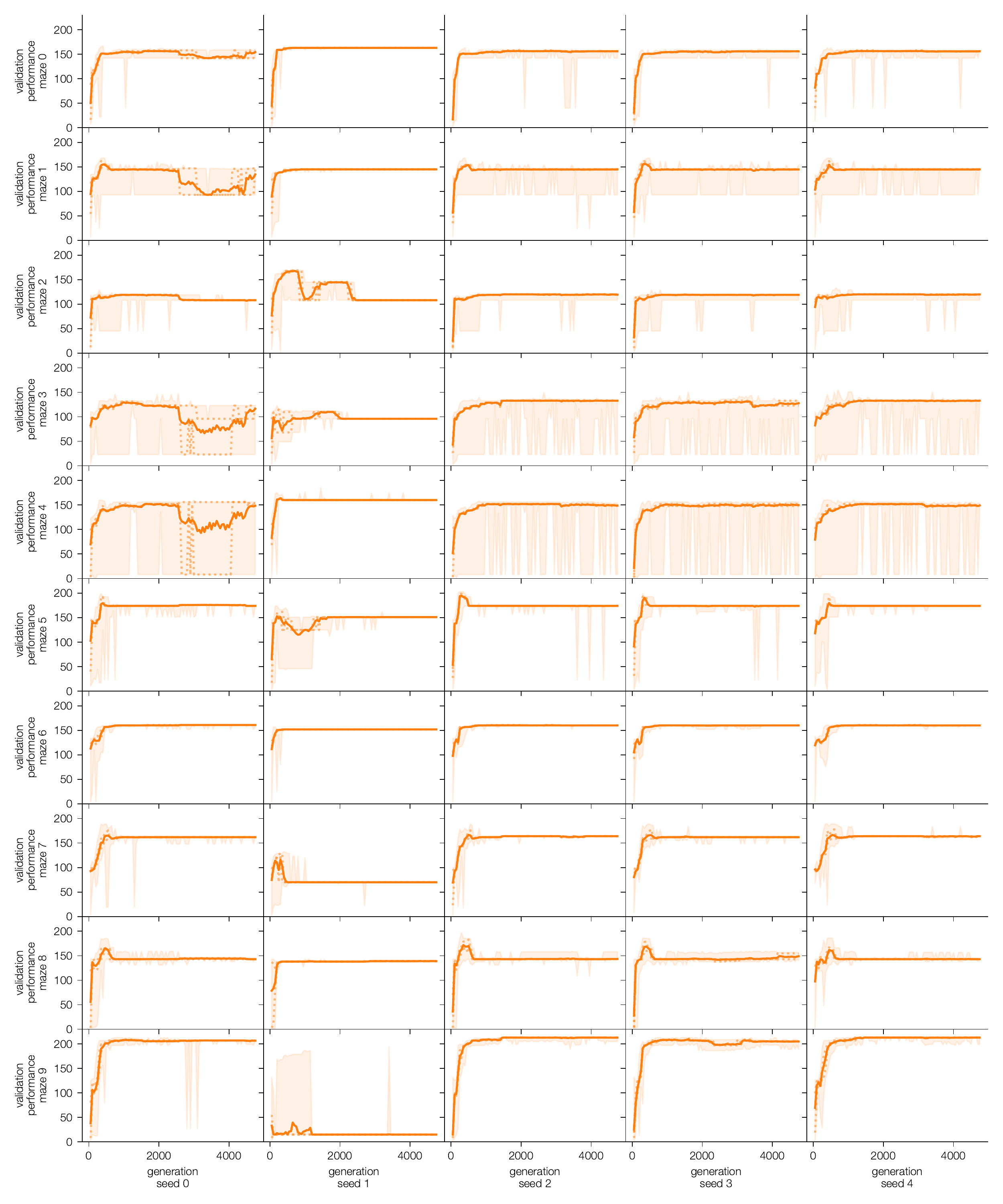}}%
\caption{\textbf{Validation performance in the "connection severance" experiment.} The validation performance over generation for different seeds (columns) and different validation mazes (rows).  Shaded area denotes the range from minimum to maximum, dotted lines indicate the 25\% and 75\% percentile, and the solid line denotes the mean.}
\label{fig:ValidationFitnessConSev}
\end{figure}

\begin{figure}[htbp]
\centering
\makebox[\textwidth][c]{\includegraphics[width=1.1\textwidth]{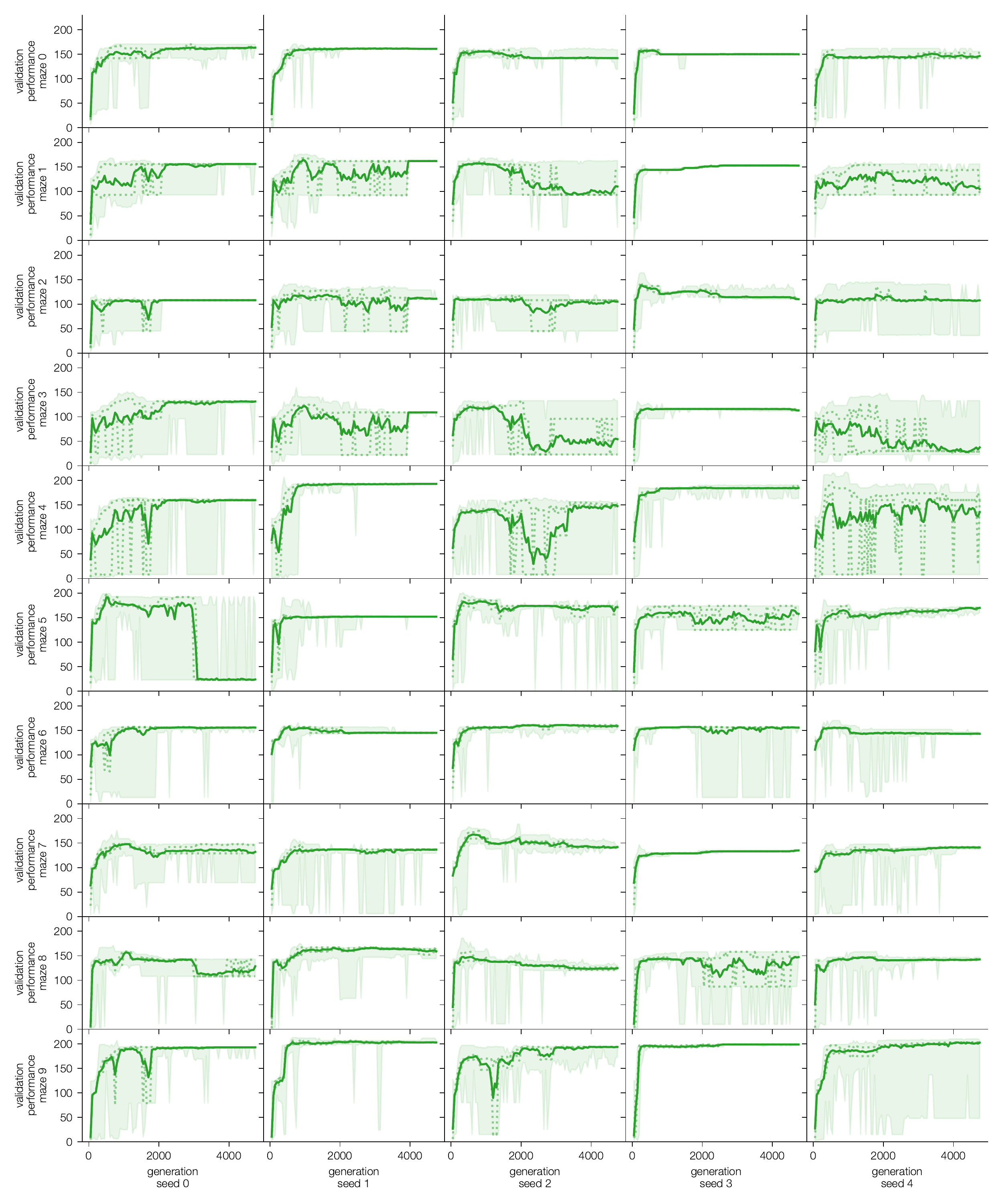}}%
\caption{\textbf{Validation performance in the "connection severance (low rate)" experiment.} The validation performance over generation for different seeds (columns) and different validation mazes (rows).  Shaded area denotes the range from minimum to maximum, dotted lines indicate the 25\% and 75\% percentile, and the solid line denotes the mean.}
\label{fig:ValidationFitnessConSevLowRate}
\end{figure}

\begin{figure}[htbp]
\centering
\makebox[\textwidth][c]{\includegraphics[width=1.1\textwidth]{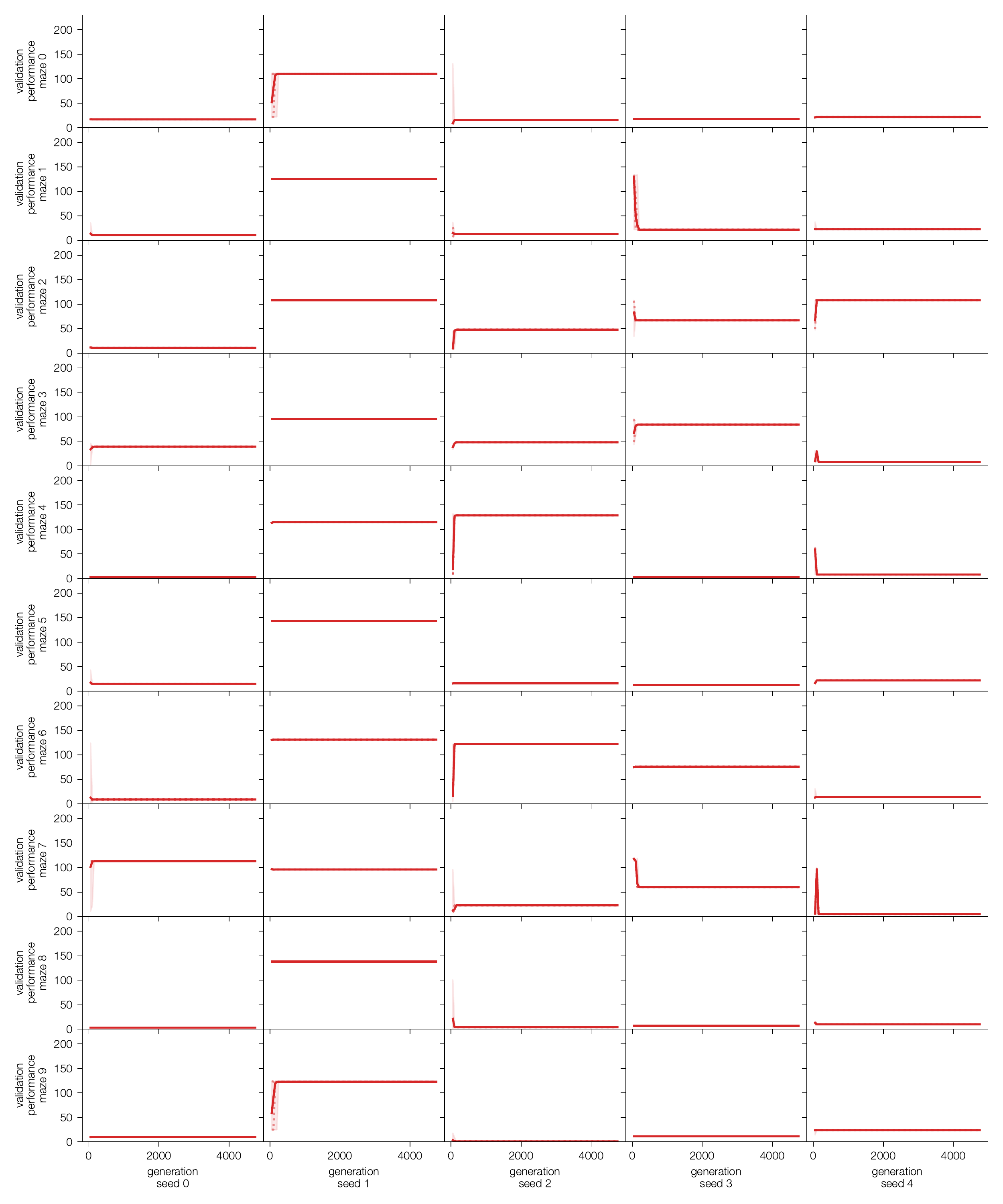}}%
\caption{\textbf{Validation performance in the "connection severance no mut" experiment.} The validation performance over generation for different seeds (columns) and different validation mazes (rows).  Shaded area denotes the range from minimum to maximum, dotted lines indicate the 25\% and 75\% percentile, and the solid line denotes the mean.}
\label{fig:ValidationFitnessConSevNoMut}
\end{figure}

\begin{figure}[htbp]
\centering
\makebox[\textwidth][c]{\includegraphics[width=1.1\textwidth]{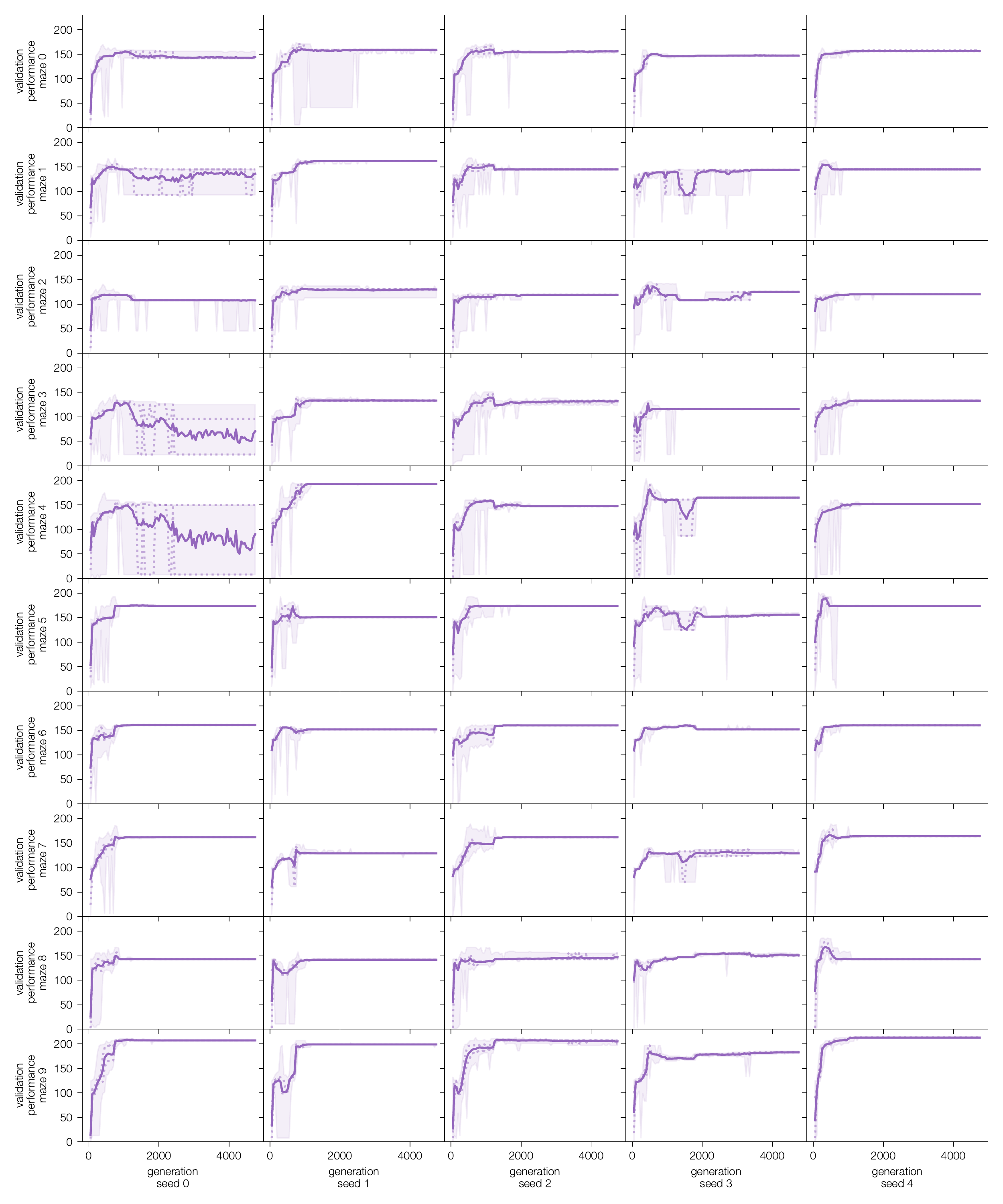}}%
\caption{\textbf{Validation performance in the "connection severance sparsity reward" experiment.} The validation performance over generation for different seeds (columns) and different validation mazes (rows).  Shaded area denotes the range from minimum to maximum, dotted lines indicate the 25\% and 75\% percentile, and the solid line denotes the mean.}
\label{fig:ValidationFitnessConSevSparsRew}
\end{figure}

\begin{figure}[htbp]
\centering
\makebox[\textwidth][c]{\includegraphics[width=1.1\textwidth]{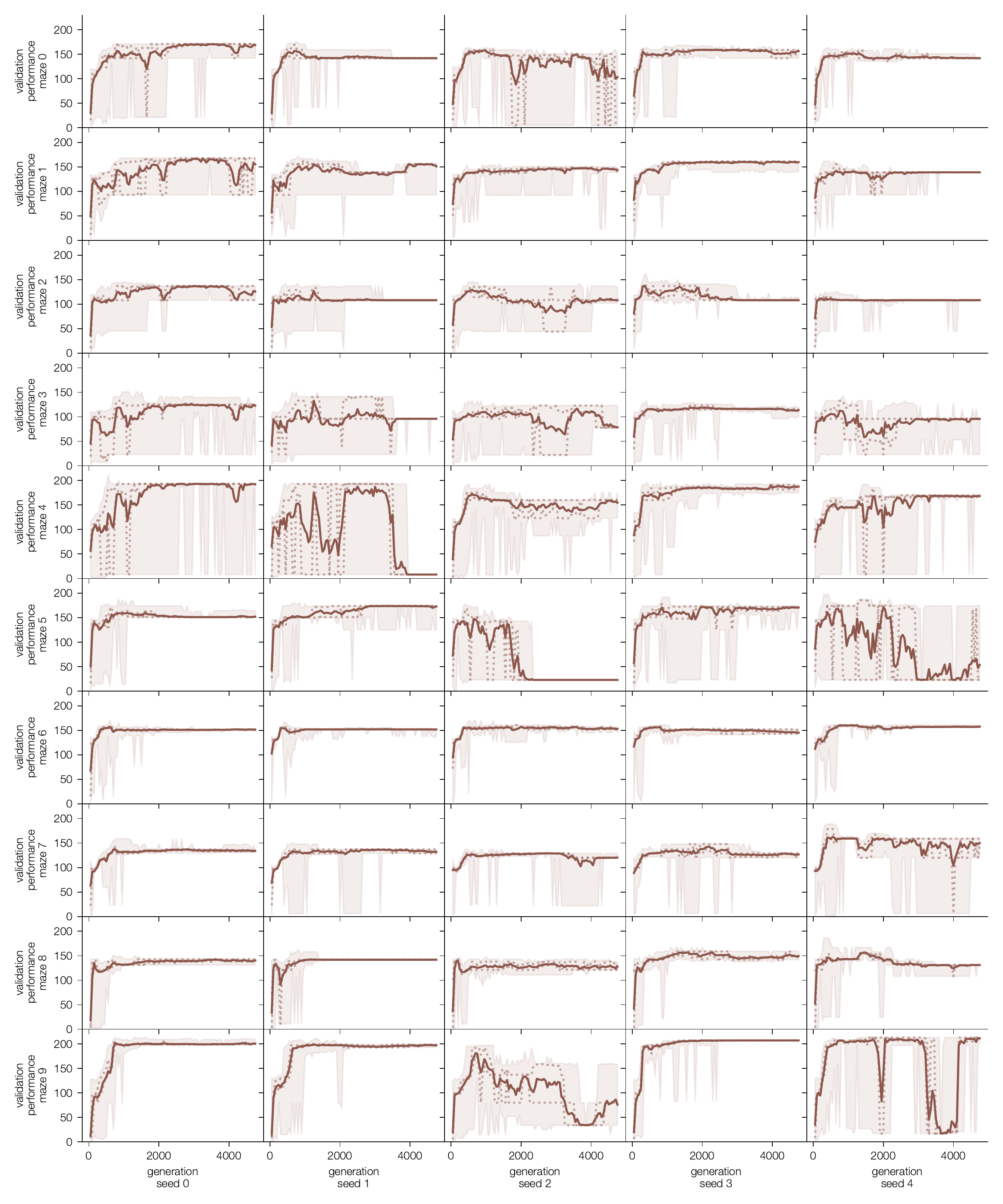}}%
\caption{\textbf{Validation performance the "sparsity reward" experiment.} The validation performance over generation for different seeds (columns) and different validation mazes (rows).  Shaded area denotes the range from minimum to maximum, dotted lines indicate the 25\% and 75\% percentile, and the solid line denotes the mean.}
\label{fig:ValidationFitnessSparsRew}
\end{figure}

\begin{figure}[htbp]
\centering
\includegraphics{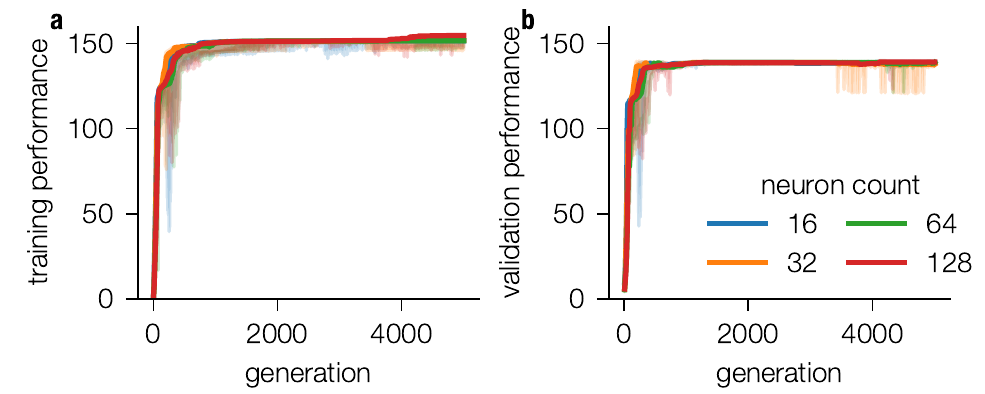}%
\caption{\hl{\textbf{Performance of the neural networks for different initialization- sizes.}  The figure shows the performance of the agents as a function of the generation for different network initialization sizes. It could be show that bigger networks do not improve the performance.}}
\label{fig:plot_network_size}
\end{figure}

\begin{figure}[htbp]
\centering
\includegraphics{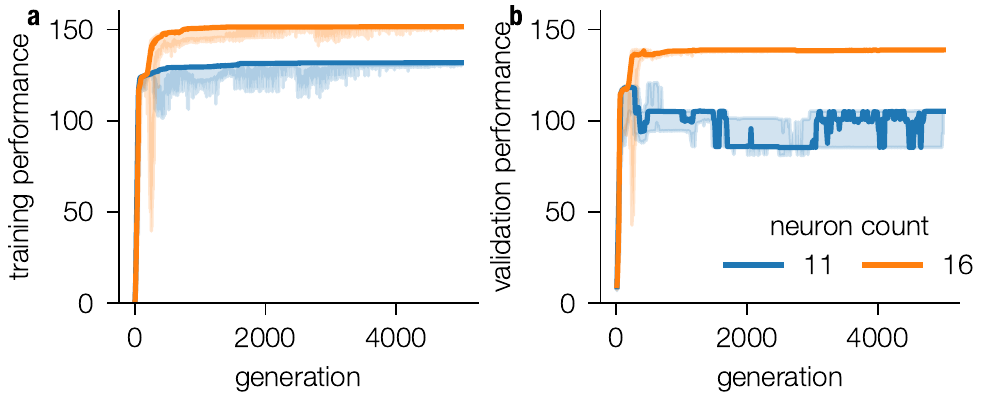}%
\caption{\hl{\textbf{Performance of the neural networks for a network with 11 neurons.}  The simulation shows that 11 neurons are not enough to fulfill the navigation task. Thus, 16 is a sophisticated network size. Thus, this size was used for our simulations.}}
\label{fig:plot_network_size_small}
\end{figure}

\begin{figure}[htbp]
\centering
\includegraphics{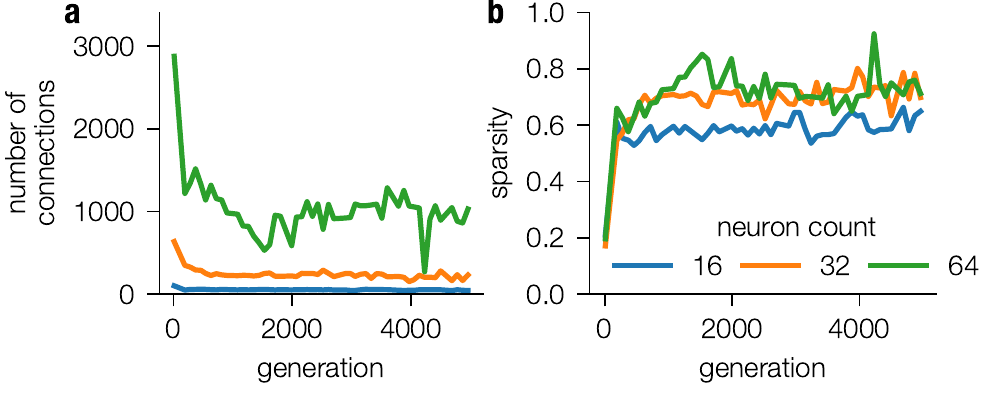}%
\caption{\hl{\textbf{Sparsity as a function of the generation for different network initialization sizes.}  The different neural networks develop equal sparsity properties.}}
\label{fig:plot_network_size_sparsity}
\end{figure}

\begin{figure}[htbp]
\centering
\includegraphics{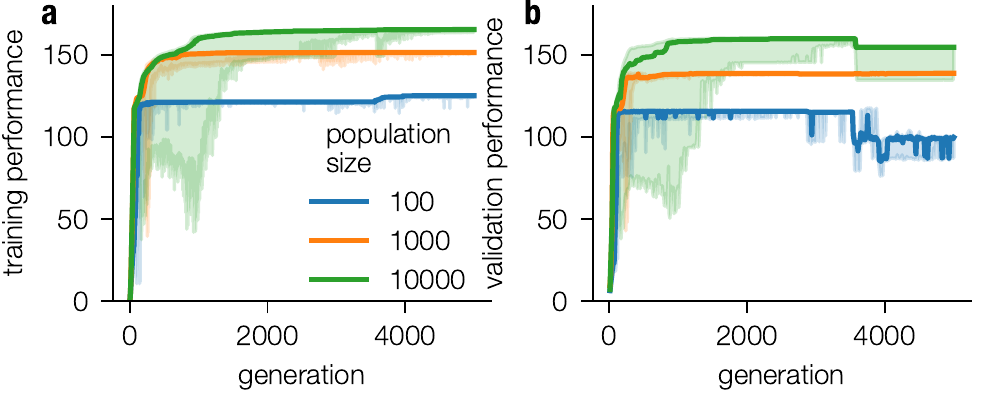}%
\caption{\hl{\textbf{Performance increase as a function of the size of the organism pool (populations size).}  The performance of the agents increases with rising pool size. Thus, a pool size of 1000 is a good trade-off of calculation speed and performance of the networks.}}
\label{fig:plot_population_size}
\end{figure}

\begin{figure}[htbp]
\centering
\includegraphics{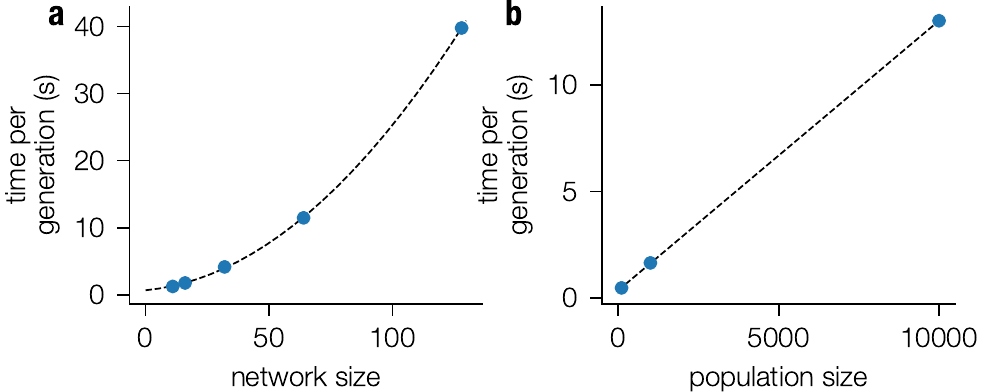}%
\caption{\hl{\textbf{Computational complexity as a function of network and population size.}  The computational complexity increases exponentially with the network size (a) and linearly with the population size (b).}}
\label{fig:plot_time_analysis}
\end{figure}

\begin{figure}[htbp]
\centering
\includegraphics{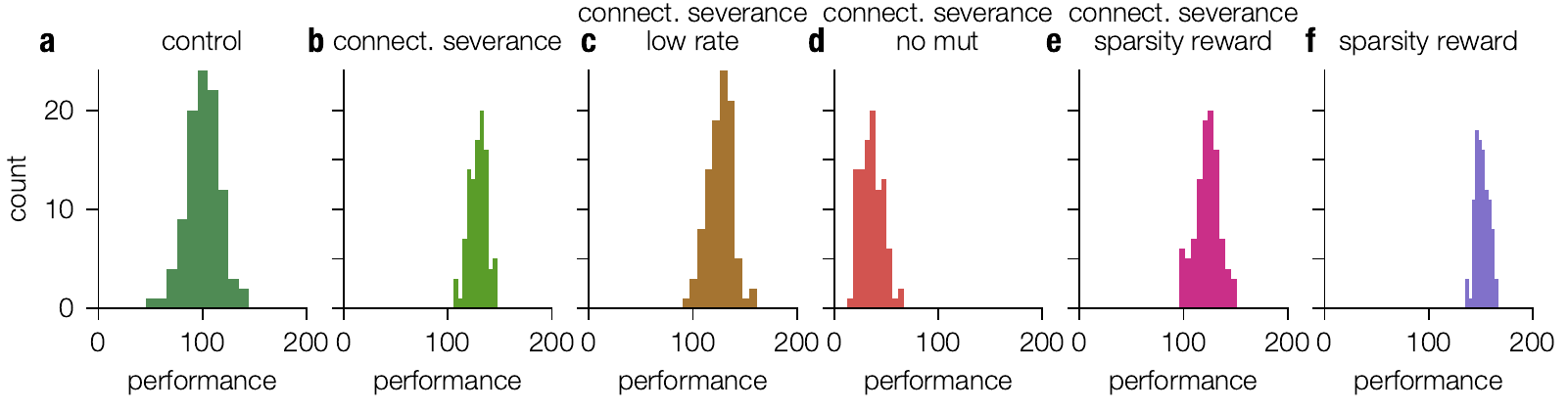}%
\caption{\hl{\textbf{Histograms of validation performance for the different sparsification approaches.}  The figure shows the histograms of the performance of the agents in 100 bunches of 10 unseen mazes each (seed 0 was used). It can clearly be seen that no sparsification causes worse performance in the unseen mazes. This emphasizes the fact that sparsification is a tool to enhance generalisation ability.}}
\label{fig:figure_crossvalidation}
\end{figure}

\begin{figure}[htbp]
\centering
\includegraphics{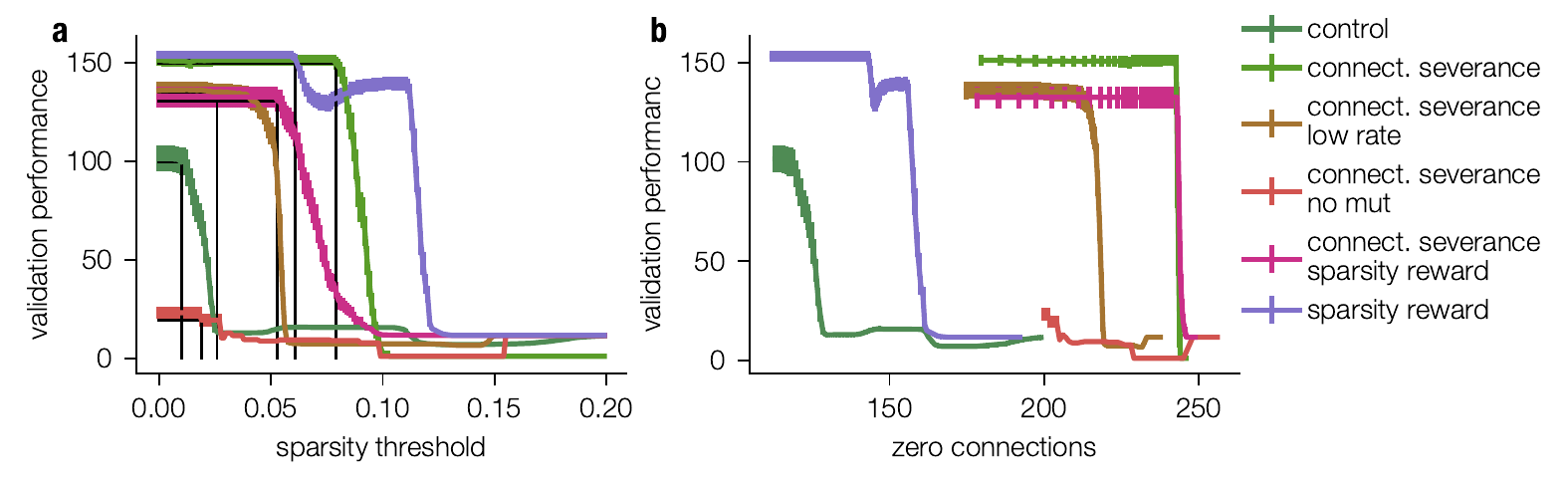}%
\caption{\hl{\textbf{Gradual deletion of network connections.} The figure shows that after evolutionary training more connections can be resolved without significantly effecting the validation performance. Thus connections, where the absolute value lies under a certain threshold are deleted and the validation performance is evaluated; a) Validation performance as function of the threshold, b) Validation performance as function of the number of connections with value 0. This shows that though the "sparsity reward" group (purple line) performs well the "connection" severance populations (light green curve) can fulfill the task with much less connections. }}
\label{fig:calc_control_to_sparse}
\end{figure}
\end{document}